\documentclass[10pt,twocolumn,letterpaper]{article}

\usepackage{cvpr}              

\usepackage{graphicx}
\usepackage{amsmath}
\usepackage{amssymb}
\usepackage{booktabs}
\usepackage{tabularx}
\usepackage{bm}
\usepackage{cite}
\usepackage{float}
\usepackage{wrapfig}
\usepackage{calc}
\usepackage[svgnames,x11names,table]{xcolor}
\usepackage{multirow}
\usepackage{tikz}
\usepackage{mdframed}
\usetikzlibrary{positioning}
\usepackage{times}
\usepackage{latexsym}
\usepackage[T1]{fontenc}
\usepackage[utf8]{inputenc}
\usepackage{microtype}
\usepackage{inconsolata}
\usepackage{listings}
\usepackage{here}
\usepackage{mathtools}
\usepackage{ulem}
\usepackage{soul}
\usepackage {colortbl,array}
\usepackage{enumitem}
\usepackage{makecell}
\usepackage{xspace}

\definecolor{TitleColor}{gray}{0.95}
\definecolor{LightCyan}{rgb}{0.88,0.95,1}
\definecolor{LightPink}{HTML}{FAE6E7}

\definecolor{att}{HTML}{A8DDA8}      \definecolor{rel}{HTML}{BFEFFF}      \definecolor{obj}{HTML}{FFAEB9}      \definecolor{num}{HTML}{FFD700}      \definecolor{txt}{HTML}{D8BFD8}      \definecolor{fact}{HTML}{98FB98}     

\newcommand{\attributetype}{\sethlcolor{att}{\hl{~Attribute~}}}
\newcommand{\relationtype}{\sethlcolor{rel}{\hl{~Relation~}}}
\newcommand{\objecttype}{\sethlcolor{obj}{\hl{~Object~}}}

\newcommand{\numbertype}{\sethlcolor{num}{\hl{~Number~}}}
\newcommand{\texttype}{\sethlcolor{txt}{\hl{~Text~}}}
\newcommand{\facttype}{\sethlcolor{fact}{\hl{~Fact~}}}

\usepackage[most]{tcolorbox}
\usepackage{cuted} \tcbuselibrary{breakable}

\newcommand{\WideTColorBox}[2]{\par\begingroup
  \onecolumn  \begin{tcolorbox}[enhanced, breakable,
    width=\textwidth,
    colback=blue!5!white,
    colframe=blue!75!black,
    fonttitle=\bfseries,
    title=#1,
    boxrule=0.5pt,
    sharp corners,
    left=2mm, right=2mm, top=1mm, bottom=1mm,
    before upper={\let\\\par},
  ]
  #2
  \end{tcolorbox}
  \endgroup
  \par
}

\newcommand{\TColorBox}[2]{\begin{tcolorbox}[enhanced, breakable,
    colback=blue!5!white,
    colframe=blue!75!black,
    fonttitle=\bfseries,
    title=#1,
    boxrule=0.5pt,
    sharp corners,
    left=2mm, right=2mm, top=1mm, bottom=1mm,
    width=\columnwidth,  before upper={\let\\\par},
  ]
  #2
  \end{tcolorbox}
}

\usepackage[pagebackref,breaklinks,colorlinks,citecolor=Blue4]{hyperref}

\usepackage[capitalize]{cleveref}
\crefname{section}{Sec.}{Secs.}
\Crefname{section}{Section}{Sections}
\Crefname{table}{Table}{Tables}
\crefname{table}{Tab.}{Tabs.}

\def\paperID{XXXX} \def\confName{CVPR}
\def\confYear{2026}

\begin{document}

\title{\textsc{Zina}: Multimodal Fine-grained Hallucination Detection and Editing}

\author{
  \begin{tabular}{cccc}
    Yuiga Wada$^{1,2,3}$ \hspace{7mm} & Kazuki Matsuda$^{2}$ \hspace{7mm} & Komei Sugiura$^{1,2}$ \hspace{7mm} & Graham Neubig$^{3}$
  \vspace{2mm}
  \end{tabular} 
  \\
  \begin{tabular}{ccc}
  $^{1}$Keio AI Research Center & $^{2}$Keio University & $^{3}$Carnegie Mellon University
  \end{tabular} \\
  \url{https://yuiga.dev/zina}
}

\maketitle
\vspace{-3mm}
\begin{abstract}
Multimodal Large Language Models (MLLMs) often generate hallucinations, where the output deviates from the visual content.
Given that these hallucinations can take diverse forms, detecting hallucinations at a fine-grained level is essential for comprehensive evaluation and analysis.
To this end, we propose a novel task of \textbf{multimodal fine-grained hallucination detection and editing} for MLLMs.
Moreover, we propose \textsc{Zina}, a novel method that identifies hallucinated spans at a fine-grained level, classifies their error types into six categories, and suggests appropriate refinements.
To train and evaluate models for this task, we construct VisionHall, a dataset comprising 6.9k outputs from twelve MLLMs manually annotated by 211 annotators, and 20k synthetic samples generated using a graph-based method that captures dependencies among error types.
We demonstrated that \textsc{Zina} outperformed existing methods, including GPT-4o and Llama-3.2, in both detection and editing tasks.

\end{abstract}

\section{Introduction}

Multimodal Large Language Models (MLLMs) have emerged as powerful systems for a broad range of vision-language tasks \cite{llama, llava-1.5, llava, gpt4, gemini, vila, qwen-vl, instructblip}. MLLMs often produce \textit{hallucinations} in image captioning tasks \cite{amber, mhaldetect, unihd},  undermining the reliability in practical applications.
To further advance the development of MLLMs, evaluating and analyzing hallucinations is essential.

\begin{figure}[t]
    \centering
    \includegraphics[width=1.0\linewidth]{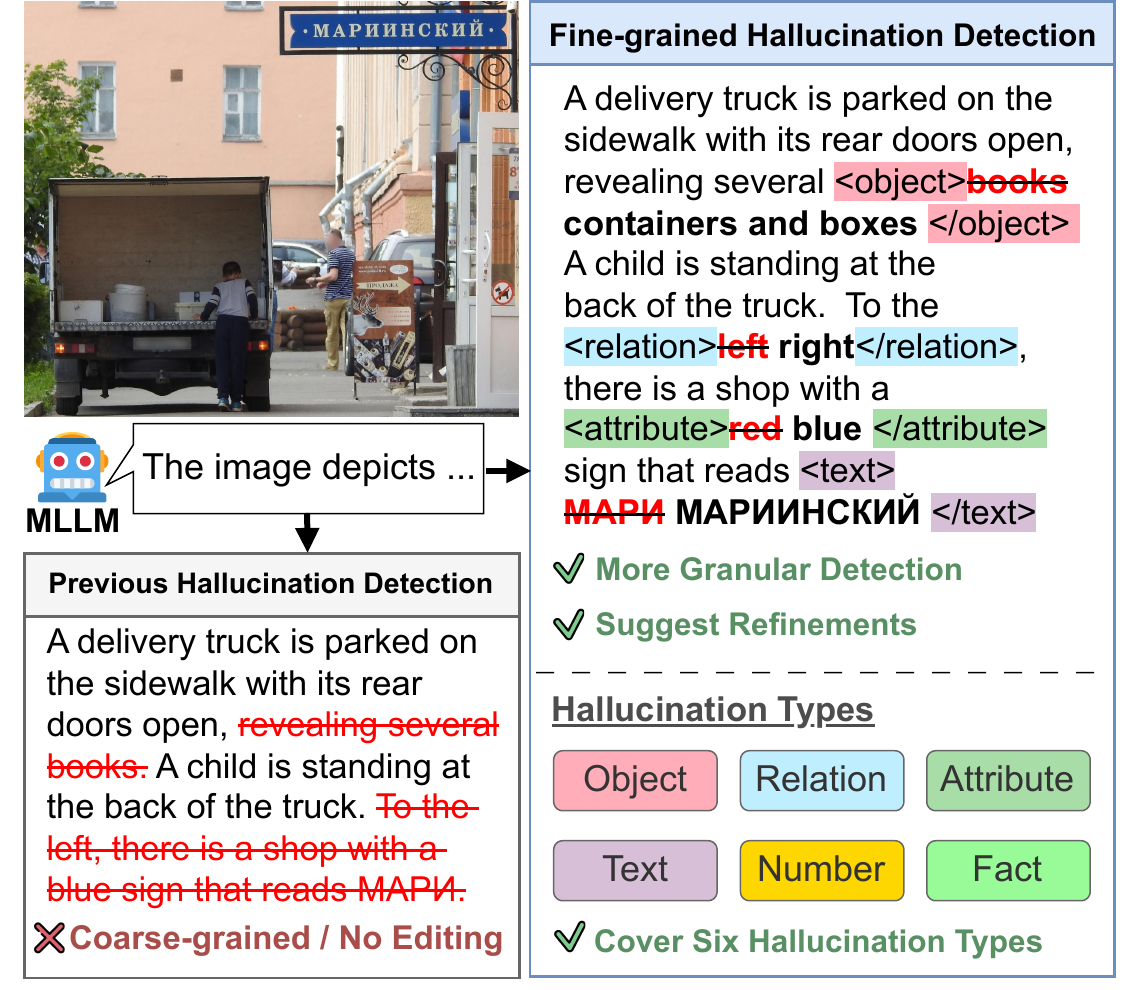}
    \caption{Overview of the proposed task. In contrast to conventional tasks, the model is expected to detect hallucinated spans at a fine-grained level, classify their types based on a taxonomy, and suggest appropriate refinements.}
    \label{fig:eye-catch}
    \vspace{-3mm}
\end{figure}

Although several works \cite{pope, mhaldetect, amber} have attempted to address hallucination detection in MLLM outputs, they have primarily focused on coarse-grained detection.
While hallucinations can take diverse forms \cite{unihd}, coarse-grained methods often classify them into only a few broad categories (e.g. only object hallucinations; \cite{pope, mhaldetect}).
Moreover, these methods typically detect hallucinations at the sentence level or across relatively long spans \cite{mhaldetect}. 

These approaches limit their ability to support detailed error analysis. For instance, consider the caption shown in Fig.~\ref{fig:eye-catch}, which contains diverse hallucinated spans~—~such as incorrect objects, colors, relationships, and scene text.
While previous methods may label phrases such as ``revealing several books'' as hallucinated, effective analysis demands pinpointing the specific error (`books'), classifying its type (`object'), and suggesting a suitable correction (``containers and boxes'')

To address these limitations, we propose \textbf{multimodal fine-grained hallucination detection and editing}, a novel task in which a model detects hallucinated spans, classifies their types according to a taxonomy, and suggests appropriate refinements.
By consolidating categories from several works~\cite{spice, unihd, fava}, our taxonomy classifies hallucinations in MLLMs' outputs into six categories.
\textit{Editing} can contribute to improving MLLMs, as they can serve as \textit{silver-standard} data, offering span-level corrections \cite{fava,mhaldetect}.

We formalize the proposed task as a tagging problem, as shown in Fig.~\ref{fig:eye-catch}.
The model is expected to place tags on each hallucinated span, similarly to previous tasks designed for text-only LLMs (e.g. \cite{fava}).
The proposed task is particularly challenging because, as we will demonstrate, even advanced MLLMs, such as GPT-4o~\cite{gpt4} and Llama-3.2~\cite{llama}, perform poorly on both the \textit{detection} and \textit{editing} tasks.

To address this challenge, we propose \textsc{Zina}, a novel method for fine-grained hallucination detection and editing in MLLMs.
The method consists of two main components: a detector MLLM that identifies hallucinated spans, and a reviewer MLLM that determines where to apply tags and suggests appropriate refinements.
Unlike previous works~\cite{fava, HalluMeasure} our method explicitly separates the responsibilities of token copying and hallucination detection/editing, reducing the complexity of each reasoning task.

Moreover, we construct the VisionHall dataset, which consists of two subsets for a detection task and an editing task.
For the detection task, the dataset contains approximately 6.9k outputs generated by twelve MLLMs, with hallucinated spans manually annotated by 211 annotators.
For the editing task, we create 20k synthetic samples by injecting errors into image captions using a novel graph-based approach, capturing the dependencies of errors.

The main contributions of this paper are as follows:
\begin{itemize}
    \setlength{\parskip}{0.2mm} \setlength{\itemsep}{0.2mm} \item We proposed a novel task of multimodal fine-grained hallucination detection and editing for MLLMs.
\item We also proposed \textsc{Zina}, a novel method designed for fine-grained hallucination detection and editing.
\item We constructed VisionHall, a semi-synthetic dataset consisting of image captions generated by 12 MLLMs.
\item \textsc{Zina} outperformed existing methods on the VisionHall dataset in both detection and editing tasks. In particular, \textsc{Zina} significantly outperformed both Llama-3.2 and GPT-4o, achieving gains of 28.2 and 15.8 points, respectively.
\end{itemize}

\section{Related Works}

\paragraph{Hallucination detection and editing in LLMs.}
Several studies have addressed hallucination detection by classifying statements based on factual correctness~\citep{manakul2023selfcheckgpt, min2023factscore, gao2022rarr, chen2023purr, li2024dawndarkempiricalstudy}.
SelfCheckGPT ~\citep{manakul2023selfcheckgpt} detects entity-level hallucinations by comparing multiple LLM outputs, treating consistent responses as factual and divergent ones as hallucinated.
\cite{li2024dawndarkempiricalstudy} employs GPT-4 to extract factual statements verifiable against world knowledge, and then identifies hallucinated spans within them. 
In contrast, \textsc{Fava}~\citep{fava} performs fine-grained hallucination detection and editing by leveraging retrieval-augmented language models.

\paragraph{Hallucination analysis in MLLMs.}
Various hallucination detection methods have been proposed for MLLMs \cite{pope, amber, mhaldetect, cceval, unihd}.
POPE \cite{pope} focuses on object hallucinations in discriminative tasks by prompting MLLMs with simple yes/no questions about specific objects.
Although POPE is limited to discriminative tasks, AMBER \cite{amber} addresses this gap by leveraging the CHAIR metric \cite{chair}, which measures the proportion of mentioned objects that are not present in the image.
Unlike these rule-based methods, MHalDetect \cite{mhaldetect} finetunes InstructBLIP and performs hallucination detection as a ternary classification task.
UniHD \cite{unihd} detects hallucinations in MLLMs by first extracting verifiable claims and then validating each claim using tools such as object detection and scene-text recognition systems.

Beyond these coarse-grained approaches, a few studies~\cite{halloc, esreal} have proposed fine-grained hallucination detection methods.
One such method is HalLocalizer~\cite{halloc}, which performs token-level localization of four hallucination types using a bidirectional VisualBERT encoder with linear classification heads. 
A key difference from these methods is that our detection task is explicitly designed for a subsequent editing task. Specifically, replacing the detected spans should be sufficient to correct the hallucination. However, existing methods~\cite{halloc, esreal, mhaldetect, unihd} such as HalLocalizer rely on token-level detection, which does not guarantee that editing only the detected tokens corrects the caption.
 
\section{Multimodal Fine-grained Hallucination Detection and Editing}
\label{sec:task}

\begin{table*}
	\centering
	\normalsize
	\vspace{-3mm}
	\setlength{\tabcolsep}{6pt}
      \begin{tabular}{llccc}
    \toprule
        Types & Example & GPT-4o [\%]&
    Q-7B [\%] &
    Q-72B [\%] \\
\midrule
        \objecttype & A \sout{dog} \textcolor{red}{cat} is sitting on the couch ... & 30.13 & 27.32 & 40.42 \\ 
        \attributetype & \sout{Blue} \textcolor{red}{Red} bicycles are leaning ... & 34.73 & 33.82 & 36.69 \\ 
        \numbertype & \sout{Three} \textcolor{red}{Four} people are sitting ... & 10.04 & 9.76 & 6.17 \\ 
        \texttype & A sign says \sout{``Restaurant''} \textcolor{red}{``Welcome''} ... & 12.27 & 14.96 & 8.60 \\ 
        \relationtype & The coffee cup is on the \sout{left} \textcolor{red}{right} side ... & 8.23 & 11.38 & 7.31 \\ 
        \facttype & \sout{Steve Jobs} \textcolor{red}{Steve Wozniak} holding  ...
        \textcolor{red}{} & 4.60 & 2.76 & 0.81 \\ 
        \bottomrule
    \end{tabular}
    \caption{Examples of hallucinations categorized by our taxonomy, along with the distribution of error types in outputs from GPT-4o, Qwen2.5-VL-7B-Instruct (Q-7B), and Qwen2.5-VL-72B-Instruct (Q-72B) {on DCI dataset~\cite{dci}.}}
    \vspace{-3mm}
	\label{table:taxonomy}
\end{table*}
We propose a novel task of \textbf{multimodal fine-grained hallucination detection and editing} for MLLMs.
We follow prior work \cite{pope, amber, mhaldetect} in defining hallucinations as errors where the MLLM-generated description deviates from the visual content of the image.

\subsection{Taxonomy}
Based on prior work \cite{spice, unihd, fava} and insights from the pilot annotation, we define a taxonomy comprising six error types:
\begin{itemize}
\setlength{\parskip}{0.2mm} \setlength{\itemsep}{0.2mm} \item \objecttype: Incorrect descriptions of specific objects or entities. \item \attributetype: Inaccurate mentions of properties such as color, size, or shape. \item \numbertype: Incorrect mentions of quantities or numerical values.
\item \texttype: Inaccurate descriptions of scene text or written content visible in the image. \item \relationtype: Errors in semantic relationships of objects (e.g., prepositions or adjectives) within the description. \item \facttype: Incorrect mentions of named entities such as people, places, or countries.
\end{itemize}
{Details of the taxonomy construction are provided in Appendix~\ref{sec:taxonomy_construction}.}

Table~\ref{table:taxonomy} presents examples of hallucinations categorized by our taxonomy.
{The table also shows the distribution of error types in generated captions for GPT-4o \cite{gpt4}, Qwen2.5-VL-7B-Instruct \cite{qwen25vl}, and Qwen2.5-VL-72B-Instruct on the DCI dataset~\cite{dci}.}
These distributions provide supporting evidence for the validity of the proposed taxonomy.

\subsection{Task Definition}
In this task, given an MLLM-generated description $x_{\mathrm{desc}}$ for an image $x_{\mathrm{img}}$, along with a reference caption $x_{\mathrm{ref}}$, a model detects hallucinated segments, classifies them according to our taxonomy, and suggests appropriate refinements.
Specifically, the output is defined as $\hat{y} = \{ \hat{\mathcal{Y}}_{\mathrm{text}}, \hat{\mathcal{Y}}_{\mathrm{edit}}, \hat{\mathcal{Y}}_{\mathrm{type}} \}$,
where $\hat{\mathcal{Y}}_{\mathrm{text}}$, $\hat{\mathcal{Y}}_{\mathrm{edit}}$, and $\hat{\mathcal{Y}}_{\mathrm{type}}$ denote the sets of hallucinated words, their corresponding edits, and their error types, respectively.
Each $\hat{y}_\mathrm{type} \in \hat{\mathcal{Y}}_{\mathrm{type}}$ is one of seven types: six from the taxonomy and one no-hallucination class.

Since reliable hallucination detection cannot be achieved using only the image as input, it is standard practice to incorporate external knowledge~\cite{fava, unihd}.  
Conventional tasks (e.g. ~\cite{unihd}) typically assume that detectors access intermediate tools such as object detectors and OCR systems to obtain such knowledge.
However, these sources often contain errors that hinder appropriate hallucination detection.
Therefore, our task assumes the availability of human-written reference captions $x_\mathrm{ref}$ as a reliable source for evaluation.

\subsection{Evaluation Metrics}
\paragraph{Detection task.}
Following previous works \cite{fava, schuster-etal-2021-get2, feng2023factkb}, we employed F$_1$ score as an evaluation metric for the detection tasks.
We first compute precision and recall by comparing 
$\{y_{\mathrm{text}}^{(i)}, y_{\mathrm{type}}^{(i)}\}_{i=1}^{N}$ and $\{\hat{y}_{\mathrm{text}}^{(i)}, \hat{y}_{\mathrm{type}}^{(i)}\}_{i=1}^{N}$ , and then calculate the F$_1$ score as the harmonic mean of precision and recall.

\vspace{-3mm}
\paragraph{Editing task.}
We evaluate editing models by sentence-level metrics, CLIP-S \cite{clipscore} and PAC-S \cite{pacs}. These metrics are suitable for evaluating edited texts as they are standard metrics for image captioning~\cite{clipscore, pacs, polos, vela, pearl}. {Further details of the evaluation metrics are provided in Appendix~\ref{appendix:metrics}.}

\vspace{-3mm}
\paragraph{Overall performance.}
Previous works (e.g. \cite{fava}) typically assessed overall model performance with sentence-level metrics such as CLIP-S~\cite{clipscore} and FactScore~\cite{min2023factscore}.
This is because edited answers often have diverse valid forms, making span-level ground truth hard to define. Sentence-level metrics evaluate editing models by computing cosine similarity between sentence embeddings. However, as they encode both edited and unedited content into a single embedding, they often overlook the impact of small edits, as unchanged portions dominate.

To address this limitation, we introduce fine-grained metrics that compare spans based on their embeddings, rather than encoding the entire sentence as a whole. Our metrics are inspired by detection metrics based on F$_1$
 scores computed via exact matching. However, instead of using exact matches, we compare spans based on embedding-based similarity.

Using the similarity function $\mathrm{sim}$, we define precision and recall as follows:
\begin{align}
&\mathrm{Precision} = \frac{\sum_{i \in N} \mathrm{sim}(\hat{y}_{\mathrm{text}}^{(i)}, y_{\mathrm{text}}^{(i)})}{\sum_{i \in N} \mathbf{1}[\hat{y}_{\mathrm{type}}^{(i)} \ne 0]}, \\
&\mathrm{Recall} = \frac{\sum_{i \in N} \mathrm{sim}(\hat{y}_{\mathrm{text}}^{(i)}, y_{\mathrm{text}}^{(i)})}{\sum_{i \in N} \mathbf{1}[y_{\mathrm{type}}^{(i)} \ne 0]},
\end{align}
where $N$ denotes the number of hallucinated words in the ground truth, respectively.
We then compute the F$_1$ score as the harmonic mean of precision and recall.
As the similarity function $\mathrm{sim}$, we adopt cosine similarity computed on BERT \cite{bert} and CLIP \cite{clip} embeddings, because their similarities are widely used as evaluation metrics in image captioning \cite{bertscore, clipscore}.
We refer to these metrics as BERT-F$_1$ and CLIP-F$_1$, respectively.

\begin{figure*}[t]
    \centering
    \includegraphics[width=1.0\linewidth]{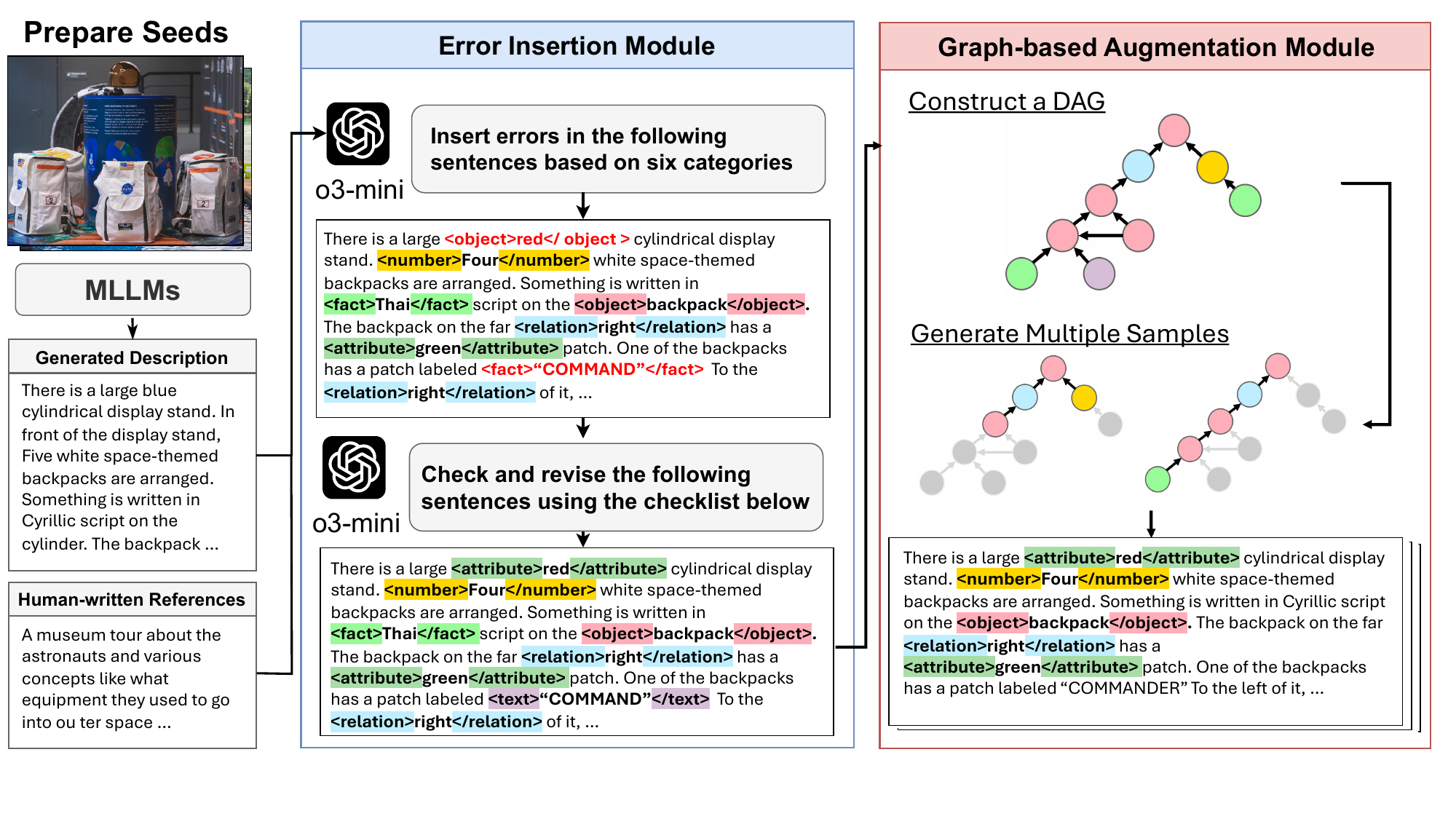}
    \caption{Overview of the graph-based synthetic data generation process. We first obtain seed descriptions by leveraging various MLLMs. Subsequently, the Error Insertion module injects errors while considering inter-span dependencies of errors. The Graph-based Augmentation module then constructs a DAG and prunes it to generate diverse training samples.}
    \label{fig:methods}
    \vspace{-3mm}
\end{figure*}

\section{Methodology}
\label{sec:methodlogy}

Previous approaches (e.g. \cite{fava}) generate outputs by copying the original sentence verbatim and inserting tags where necessary.  
These approaches face several challenges:  
(i) the model should reproduce the original sentence exactly, word by word;  
(ii) these approaches require a model to simultaneously determine, token by token, where to insert opening and closing tags;  
(iii) due to exposure bias in autoregressive generation, a single mistake can compromise the structural consistency of the entire output, often resulting in malformed tag sequences.

\subsection{Proposed Method: \textbf{\textsc{Zina}}}
To address these limitations, we propose \textsc{Zina}, a two-stage system that detects hallucinations based on our taxonomy and suggests refinements.
The proposed method consists of two main components: the detector MLLM $\mathcal{M}_\mathrm{det}$ and the reviewer MLLM $\mathcal{M}_\mathrm{rev}$.
Our central strategy is to \textit{delegate token copying to a deterministic function}, allowing the language model to focus solely on the detection and editing sub-tasks for erroneous spans.
This strategy, which decouples the detection and tagging processes, can be broadly applied to existing hallucination detection methods \cite{fava, mhaldetect}.

We begin by constructing the prompt $p_\mathrm{det}$ using $\{ x_{\mathrm{desc}},\ x_{\mathrm{img}},\ x_{\mathrm{ref}} \}$.  
In few-shot settings, $p_\mathrm{det}$ additionally incorporates $n$ few-shot examples.  
The prompt design follows prior work (e.g., \cite{fava}), and the full prompt templates are provided in the Appendix \ref{appendix:prompts}.  
We then feed $p_\mathrm{det}$ into the detector MLLM $\mathcal{M}_\mathrm{det}$ to obtain hallucinated spans $\{h_{\text{type}}^{(i)}\}_{i=1}^{M}$ along with their corresponding error types $\{h_{\text{error}}^{(i)}\}_{i=1}^{M}$, where $M$ denotes the number of predicted hallucinations.

Subsequently, for each hallucinated span, we generate a tagged sequence $z_i$ as follows:
\begin{align}
z_i = \mathcal{T}(x_{\text{desc}}, h_{\text{text}}^{(i)}, h_{\text{type}}^{(i)}),
\end{align}
where $\mathcal{T}$ is a deterministic function that inserts tags into the hallucinated span based on its error type.
Unlike previous works \cite{fava}, we adopt $\mathcal{T}$ as it delegates token copying to a deterministic function. This design allows $\mathcal{M}_\mathrm{det}$ to focus exclusively on the hallucination detection.

Given $\{z_i\}_{i=1}^M$, the reviewer MLLM $\mathcal{M}_{\text{rev}}$ assesses whether the hallucinated span is appropriately tagged within its surrounding context, and then outputs $ \hat{\mathcal{Y}}_{\mathrm{text}}$ and $ \hat{\mathcal{Y}}_{\mathrm{type} }$ as follows:
\begin{align}
\hat{y} = \mathcal{M}_\mathrm{rev}(z_i, x_\mathrm{img}, x_\mathrm{ref}).
\end{align}
In the editing setting, $\mathcal{M}_{\text{rev}}$ additionally generates suitable refinements $\hat{\mathcal{Y}}_{\mathrm{edit}}$ for each hallucinated span.
We use Qwen2.5-VL-72B-Instruct \cite{qwen25vl} to initialize $\mathcal{M}_\mathrm{det}$ and $\mathcal{M}_\mathrm{rev}$.  
For both models, we employ cross-entropy as the loss function.

\subsection{Synthetic Training Data Curation}
\label{sec:data-curation}
The development of a hallucination editor requires a diverse set of training samples.
In this field, it is standard to synthesize training data by injecting artificial hallucinations into correct sentences \cite{asai2023self,fava,halloc}.
These methods assume hallucinations are self-contained and span-localized; however, image-grounded hallucinations in MLLMs often span multiple semantically related regions, which makes it essential to capture dependency structures among errors.

To this end, we propose a novel data construction method that explicitly handles error dependencies using graph structures.
Our approach consists of two main modules: the Error Insertion (EI) module and the Graph-based Augmentation (GraphAug) module.
Fig.~\ref{fig:methods} shows an overview of our data generation pipeline.
We first obtain seed descriptions. The EI module then injects errors while considering dependencies, and the GraphAug module constructs a graph and prunes it for generating diverse errors.

\vspace{-3mm}
\paragraph{Seed descriptions.} 
In general, synthetic data construction requires a set of seed descriptions as input. Prior work \cite{fava} generated seed data from external knowledge sources that are guaranteed to be factually correct. Synthetic hallucinations are then introduced by replacing specific spans within these descriptions.
However, in image-grounded generative tasks, factually guaranteed descriptions are scarce, making such approaches less applicable.
Although standard image captioning datasets provide reference captions \cite{coco, nocaps}, naively replacing parts of these captions tends to produce limited variation and may be suboptimal for training $\mathcal{M}_{\text{rev}}$.

Therefore, we use MLLM-generated descriptions that were judged by annotators to be free of hallucinations as seed inputs.
Specifically, we leverage hallucination-free descriptions from the newly constructed VisionHall dataset, as detailed in Section~\ref{sec:visionhall}.

\vspace{-3mm}
\paragraph{Error insertion.} 
The EI module first injects hallucinations into the hallucination-free descriptions while simultaneously capturing the dependencies of errors.
To perform the insertion, we adopt o3-mini \cite{o3}, which enables simultaneously performing hallucination injection and explicit capture of inter-error dependencies.
This module outputs data in XML format, which allows for the specification of hierarchical dependencies.
{Here, we define an inter-error dependency as a causal or referential relationship between two or more hallucinated spans, where one hallucination (e.g., a non-existent object) induces or conditions the generation of subsequent hallucinations (e.g., attributes or relations related to that object). For example, if an MLLM mentions ``apples'' in the initial sentence despite no apples being present in the image, it may then mention incorrect relations between the apples and other objects due to its autoregressive nature, as pointed out in \cite{zhong-etal-2024-investigating}.}

After insertion, the module re-validates the dependencies for consistency, as we empirically observed that inserted errors often link to non-existent spans.
The full prompt used for this process is provided in Appendix \ref{appendix:prompts}.

\vspace{-3mm}
\paragraph{Graph-based augmentation.} 
The GraphAug module builds a directed graph representing dependencies among inserted errors. It removes cycles by detecting them and deleting descendant nodes, yielding a Directed Acyclic Graph (DAG) that captures structural error relationships.  

Subsequently, the module prunes the DAG by randomly selecting a node with probability $p$ and removing it along with its descendants. This enables the generation of diverse training samples with varying types and combinations of hallucinations for effective editing model training.

\section{VisionHall Dataset}
\label{sec:visionhall}

\begin{table}[t]
  \centering
  \setlength{\tabcolsep}{6pt}
  \begin{tabular}{lcccc}
    \toprule
    \multirow{2}{*}{Category} &
      \multicolumn{2}{c}{\shortstack{Avg.\ Length\\\relax [words]}} &
      \multicolumn{2}{c}{\shortstack{Frequency\\\relax [\%]}} \\
    \cmidrule(lr){2-3}\cmidrule(lr){4-5}
    & Real & Synthetic & Real & Synthetic \\
    \midrule
    \objecttype    & 1.47  & 1.15  & 41.57 & 37.41 \\
    \facttype      & 2.99 & 1.77 & 4.55  & 10.41 \\
    \texttype      & 1.79 & 1.50  & 3.21  & 7.84  \\
    \numbertype    & 1.18  &  1.07  & 10.51 & 15.40 \\
    \relationtype  & 1.78  &  1.25  & 9.06  & 13.12 \\
    \attributetype & 1.32  & 1.01  & 31.10 & 15.81 \\
    \bottomrule
  \end{tabular}

\caption{Comparison between real and synthetic samples in terms of span lengths and error frequencies. The average lengths and frequencies are generally similar across real and synthetic data.}
  \label{tab:len_freq}
\vspace{-3mm}
\end{table}

\begin{table*}[t]
	\centering
	\normalsize
\setlength{\tabcolsep}{6pt}
	\begin{tabular}{lccccc}
		\toprule
        \multirow{2}{*}{\textbf{Method}}
		& \multicolumn{1}{c}{\textbf{Detection}}
		& \multicolumn{2}{c}{\textbf{Editing}}
		& \multicolumn{2}{c}{\textbf{Overall}} \\
		\cmidrule(lr){2-2}\cmidrule(lr){3-4}\cmidrule(lr){5-6}
		&
		\textbf{F$_1$}
		& \textbf{CLIP-S} & \textbf{PAC-S}
		& \textbf{BERT-F$_1$} & \textbf{CLIP-F$_1$} \\
		\midrule

		LLaVA-1.5-7B~\cite{llava-1.5}                  & 0.82        & 64.01             & 72.72        & 0.66              & 0.93                         \\
		Qwen2-VL-7B~\cite{qwen2vl}                   & 3.36            & 64.79             & 73.01     & 3.62              & 4.98                        \\
		LLaVA-OV-Qwen2-7B~\cite{llavaov}             & 3.39          & 64.06             & 72.40     & 3.39              & 3.39                          \\
		LLaVA-1.5-13B~\cite{llava-1.5}                & 4.73           & 64.74             & 73.02     & 5.08              & 6.71                         \\
		LLaVA-NeXT-Qwen-32B~\cite{llavanext}           & 19.09       & 65.34             & 73.47         & 24.29             & \underline{31.06}           \\
		Llama-3.2-90B-Vision-Instruct~\cite{llama} & 16.92       & 65.28             & 73.54        & 14.56             & 17.62                        \\
		Qwen2.5-VL-72B-Instruct ~\cite{qwen25vl}      & 21.31        & 64.38             & 72.99        & 18.85             & 23.67                       \\
		LLaVA-OV-Qwen2-72B~\cite{llavaov}            & 25.70       & \underline{65.74} & 73.91         & 20.81             & 26.81                       \\
		GPT-4o (w/o images)~\cite{gpt4}           & 27.02         & 65.66             & \underline{73.99}     & 23.34             & 27.99             \\
		GPT-4o~\cite{gpt4}                        & \underline{29.37} & 65.58             & 73.86  & \underline{24.89} & 30.19                         \\
		\rowcolor{LightPink}
		{} & 
		\textbf{45.15} & \textbf{66.08} & \textbf{74.36}  & \textbf{44.02} & \textbf{50.39} \\
		\rowcolor{LightPink}
		\multirow{-2}{*}{\textbf{\textsc{Zina} (Ours)}}
		& (\textcolor{blue}{+15.8}) & (\textcolor{blue}{+0.34}) & (\textcolor{blue}{+0.37}) & (\textcolor{blue}{+19.1}) & (\textcolor{blue}{+20.2})
		 \\
		\bottomrule
	\end{tabular}
        \caption{Quantitative comparison with baseline methods on the VisionHall dataset. \textbf{Bold} font indicates the best, and \underline{underlined} font indicates the second best. Our proposed methods outperformed the baselines in both tasks.}
	\label{table:quant}
\vspace{-3mm}
\end{table*}

We constructed VisionHall, a dataset specifically designed to train and evaluate the fine-grained hallucination detector and editor.
Existing datasets for hallucination detection, such as AMBER~\cite{amber} and M-HalDetect \cite{mhaldetect}, are limited to coarse-grained labels (e.g. binary or ternary) and thus unsuitable for evaluating fine-grained detectors.
Furthermore, these datasets (e.g. \cite{unihd}) lack ground-truth refinements for hallucinated spans, making them unsuitable for editing tasks.

In contrast, a few datasets address fine-grained hallucination detection (e.g. \cite{fava}).
However, they mainly focus on factual errors or unverifiable statements given world knowledge, and do not cover image-grounded hallucinations where the description deviates from the visual content.

To address these limitations, we constructed VisionHall, a dataset that enables comprehensive evaluation of fine-grained hallucination detection and editing.
Each sample also includes the corresponding image and a human-written long reference.

Since human-written references offer a reliable basis for evaluation, reference-based evaluation is standard practice in natural language generation tasks \cite{bertscore, bartscore, polos}.
Accordingly, we utilize the reference captions provided in the DCI dataset \cite{dci}.
These captions are comprehensive, human-written descriptions that cover nearly all elements in each image.

As the DCI dataset provides only image–reference pairs, we collected MLLM-generated descriptions from twelve representative MLLMs and image captioning models: GPT-4o~\cite{gpt4}, 
Qwen2.5-VL 7B~\cite{qwen25vl}, Qwen2.5-VL 72B, LLaVA-NeXT~\cite{llavanext}, LLaVA-1.5~\cite{llava-1.5}, MultimodalGPT~\cite{mmgpt}, Qwen-VL-Chat~\cite{qwen-vl}, ShareGPT4V~\cite{sharegpt4v}, InstructBLIP~\cite{instructblip}, InternVL~\cite{internvl}, BLIP2~\cite{blip2}, and  GIT~\cite{git}. 
We employed the official prompts for each model to generate these outputs.

Subsequently, human annotators labeled hallucinated spans in the generated descriptions according to our taxonomy.
The annotation was conducted via a crowdsourcing platform, and responses exhibiting suspicious behavior were excluded to preserve data quality.
Full annotation guidelines and interface are provided in the Appendix~\ref{sec:construction}.

For the detection task, VisionHall comprises 6,854 MLLM-generated descriptions for 4,759 images, collected from 211 annotators.
For the editing task, our dataset contains 20k synthetic samples by our novel graph-based method, as detailed in Section \ref{sec:data-curation}.
{
Further details on the VisionHall dataset are provided in Appendix~\ref{sec:construction} and \ref{sec:ablation_curation}.
}

{
Table~\ref{tab:len_freq} shows the comparison between real and synthetic samples in terms of span lengths and error frequencies.
For most categories, the span lengths between real and synthetic samples are generally comparable.
The overall distributions are also similar across real and synthetic samples.
Further details of comparisons between real and synthetic samples are provided in Appendix~\ref{sec:statistics}.
}

\begin{table*}[t]
\centering
\normalsize
\setlength{\tabcolsep}{5pt}
\begin{tabular}{>{\centering\arraybackslash}m{3mm}lcccccccccc}
\toprule
& \multirow{2}{*}{\textbf{Model}} &
\multicolumn{3}{c}{\textbf{Hallucinatory}} &
\multicolumn{3}{c}{\textbf{Non-Hallucinatory}} &
\multicolumn{4}{c}{\textbf{ Average}} \\
\cmidrule(lr){3-5}\cmidrule(lr){6-8}\cmidrule(lr){9-12}
& & \textbf{P} & \textbf{R} & \textbf{F$_1$}
& \textbf{P} & \textbf{R} & \textbf{F$_1$}
& \textbf{Acc.} & \textbf{P} & \textbf{R} & \textbf{Mac.F$_1$} \\
\midrule
\multirow{7}{*}{\rotatebox[origin=c]{90}{\textbf{Claim-level}}}
& Gemini-based Self-Check ($n=0$) & 83.17 & 42.15 & 55.95 & 55.64 & \underline{89.48} & 68.61 & 63.34 & 69.41 & 65.82 & 62.28 \\
& Gemini-based Self-Check ($n=2$) & 84.24 & 66.75 & 74.48 & 67.35 & 84.60 & 75.00 & 74.74 & 75.80 & 75.68 & 74.74 \\
& GPT-based Self-Check ($n=0$) & 84.78 & 80.07 & 82.35 & 61.64 & 69.01 & 65.12 & 76.56 & 73.21 & 74.54 & 73.73 \\
& GPT-based Self-Check ($n=2$) & 86.54 & 85.13 & \underline{85.83} & 69.05 & 71.48 & 70.24 & 80.80 & 77.80 & 78.30 & 78.04 \\
& Gemini-based UniHD & \underline{84.44} & 72.44 & 77.98 & 71.08 & 83.54 & 76.80 & 77.41 & 77.76 & 77.99 & 77.39 \\
& GPT-based UniHD & 82.54 & \underline{85.29} & {83.89} & \underline{81.08} & 77.74 & \underline{79.38} & \underline{81.91} & \underline{81.81} & \textbf{81.52} & \underline{81.63} \\
\rowcolor{LightPink}
& \textbf{\textsc{Zina} (Ours)} & \textbf{84.91} & \textbf{89.52} & \textbf{87.15} & \textbf{84.91} & \textbf{89.52} & \textbf{87.15} & \textbf{85.39} & \textbf{86.07} & \underline{80.28} & \textbf{83.07} \\
\midrule
\multirow{7}{*}{\rotatebox[origin=c]{90}{\textbf{\;Segment-level}}}
& Gemini-based Self-Check ($n=0$) & 89.30 & 47.71 & 62.19 & 43.76 & \textbf{87.68} & 58.38 & 60.38 & 66.53 & 67.69 & 60.29 \\
& Gemini-based Self-Check ($n=2$) & \textbf{90.44} & 71.08 & 79.60 & 57.35 & \underline{83.80} & 68.10 & 75.11 & 73.89 & 77.44 & 73.85 \\
& GPT-based Self-Check ($n=0$) & 79.37 & 74.17 & 76.68 & 70.52 & 76.22 & 73.26 & 75.09 & 74.94 & 75.19 & 74.97 \\
& GPT-based Self-Check ($n=2$) & 82.00 & 79.98 & 80.98 & 76.04 & 78.35 & \underline{77.18} & 79.25 & 79.02 & 79.16 & 79.08 \\
& Gemini-based UniHD & 88.77 & 78.76 & 83.46 & 63.17 & 78.52 & 70.02 & 78.68 & 75.97 & 78.64 & 76.74 \\
& GPT-based UniHD & 87.03 & \underline{91.01} & \underline{88.98} & \underline{78.52} & 70.77 & {74.44} & \underline{84.60} & \underline{82.77} & \underline{80.89} & \underline{81.71} \\
\rowcolor{LightPink}
& \textbf{\textsc{Zina} (Ours)} & \underline{89.53} & \textbf{93.47} & \textbf{91.46} & \textbf{84.38} & 76.33 & \textbf{80.15} & \textbf{88.06} & \textbf{86.95} & \textbf{84.90} & \textbf{85.80} \\
\bottomrule
\end{tabular}
\vspace{-3mm}
\caption{Quantitative results on the ``Image-to-Text'' subset of the MHaluBench dataset. F$_1$ denotes the Micro-F$_1$ score, and Mac-F$_1$ represents the Macro-F$_1$ score. $n$ denotes the number of examples used in the few-shot setting. \textbf{Bold} indicates the best and \underline{underlined} indicates the second best. 
We employed the same baseline models as Chen et al~\cite{unihd}.}
\label{tab:mhalubench}
\vspace{-3mm}
\end{table*}

\begin{figure*}[t]
    \centering
      \includegraphics[width=1.0\linewidth]{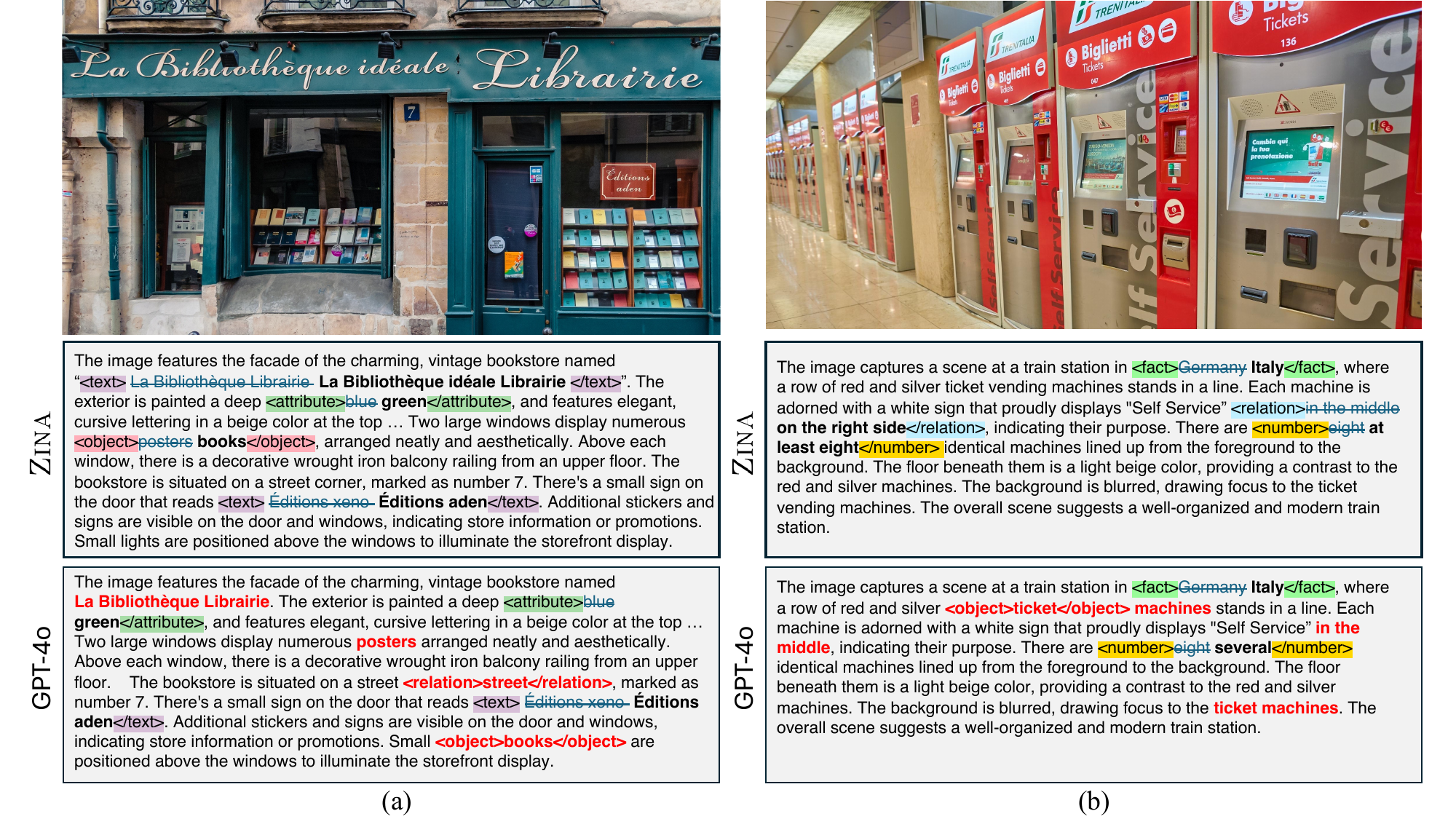}
    \caption{Qualitative results on the VisionHall dataset. Each subfigure shows the image and the edited descriptions generated by GPT-4o and \textsc{Zina}. Edited spans are enclosed in tags; strikethrough text indicates the original hallucinated phrase, while \textbf{bold} text shows the suggested refinement. \textcolor{red}{Red} spans indicate errors such as incorrect tagging or missed detections.}
    \label{fig:results}
\vspace{-5mm}
\end{figure*}

\vspace{-1mm}
\section{Experiments}

\subsection{Setup}
\paragraph{Datasets.}

To assess the practicality of the methods, it is essential to evaluate them on both in-domain and out-of-domain datasets.
However, to the best of our knowledge, there is no publicly available dataset for our proposed tasks.
Therefore, in addition to the VisionHall dataset, we employed the MHaluBench~\cite{unihd} dataset to evaluate performance on the coarse-grained hallucination detection task.
This dataset provides samples with coarse-grained hallucination annotations in both Image-to-Text and Text-to-Image settings.
We evaluated the baselines and our proposed method on the ``Image-to-Text'' subset of the MHaluBench dataset because our task focused on image captioning.

\paragraph{Baselines.}
We adopted the following MLLMs as baselines on the VisionHall dataset: LLaVA-1.5-7B~\cite{llava-1.5}, LLaVA-1.5-13B, Qwen2-VL-7B~\cite{qwen2vl}, LLaVA-OV-Qwen2-7B~\cite{llavaov}, LLaVA-NeXT-Qwen-32B~\cite{llavanext}, LLaVA-OV-Qwen2-72B, Llama-3.2-90B-Vision-Instruct~\cite{llama}, Qwen2.5-VL-72B-Instruct~\cite{qwen25vl}, and GPT-4o~\cite{gpt4}.
These models were selected as they are standard and representative MLLMs. 
A modified version of the FAVA prompt~\cite{fava} with a 3-shot setting was used for all evaluations on VisionHall. 
For the evaluation on MHaluBench, we employed the same baseline models as Chen et al~\cite{unihd}.
Implementation details and prompts are provided in Appendices \ref{appendix:impl} and \ref{appendix:prompts}.
\subsection{Results}

\paragraph{Quantitative results.}
Table~\ref{table:quant} presents a quantitative comparison with baseline methods on the VisionHall dataset.
Our proposed method achieved an F$_1$ score of 45.15 in the detection task.
Moreover, for the editing task, it achieved scores of 44.02, 50.39, 66.08, and 74.36 on BERT-F$_1$, CLIP-F$_1$, CLIP-S, and PAC-S, respectively.
These results demonstrate that our method outperformed baseline models by 15.78 points on F$_1$, and by 19.14, 19.33, 0.34, and 0.37 points on BERT-F$_1$, CLIP-F$_1$, CLIP-S, and PAC-S, respectively.

Table~\ref{tab:mhalubench} shows the quantitative results on the ``Image-to-Text'' subset of the MHaluBench dataset.
Following Chen et al.~\cite{unihd}, we reported the evaluation results at both the segment-level and the claim-level.
As a result, our proposed method outperformed the baselines on 9 out of 10 metrics at the claim-level and on 8 out of 10 metrics at the segment-level.
These results indicate that \textsc{Zina} also demonstrated strong performance on out-of-domain data.
Further quantitative results are provided in the Appendix.

\begin{table*}[t]
\centering
\normalsize
\vspace{-3mm}
\setlength{\tabcolsep}{6pt}
\begin{tabular}{lccccccccc}
\toprule
\multirow{2}{*}{\textbf{Model}}
& \multirow{2}{*}{$\boldsymbol{\mathcal{M}_{\text{rev}}}$}
& \multirow{2}{*}{\textbf{Backbone}}
& \multirow{2}{*}{$\bm{n}$}
& \multicolumn{1}{c}{\textbf{Detection}}
& \multicolumn{2}{c}{\textbf{Editing}} 
& \multicolumn{2}{c}{\textbf{Overall}} \\
\cmidrule(lr){5-5}\cmidrule(lr){6-7}\cmidrule(lr){8-9}
& & & & \textbf{F$_1$}
& \textbf{CLIP-S} & \textbf{PAC-S}
& \textbf{BERT-F$_1$}     & \textbf{CLIP-F$_1$} \\
\midrule
(i) &        & Qwen2.5-VL-72B   & 3 & 21.91 & 63.32 & 71.69 & 15.54 & 17.88  \\
(ii) &        $\checkmark$       & Qwen2.5-VL-32B    & 3 & 32.55 & 61.20 & 72.63 & 27.52 & 34.66  \\
(iii) & $\checkmark$ & LLaVA-OV-72B     & 3 & 34.41& 65.78 & 74.02 & 31.39 & 36.10  \\
(iv) & $\checkmark$ & Qwen2.5-VL-72B   & 1 & 43.25 & 65.18 & 73.30 & 42.53 & 49.54  \\
(v) & $\checkmark$ & Qwen2.5-VL-72B   & 2 & 44.21 & 65.99 & 74.32 & 43.39 & 50.21  \\
\rowcolor{LightPink}
(vi) & $\checkmark$ & Qwen2.5-VL-72B   & 3 &
\textbf{45.15} & \textbf{44.02} & \textbf{50.39} & \textbf{66.08} & \textbf{74.36} \\
\bottomrule
\end{tabular}
\caption{Ablation study of the proposed method. $\mathcal{M}_\mathrm{rev}$ indicates whether the reviewer LLM is used in addition to the detector, and $n$ denotes the number of examples used in the few-shot setting.}
\label{table:ablation}
\vspace{-2mm}
\end{table*}

\paragraph{Qualitative results.}
Fig.~\ref{fig:results} shows examples from the VisionHall dataset.
In the example on the left, the original description included hallucinations of the \objecttype, \attributetype, and \texttype types.
GPT-4o failed to detect scene text errors related to the names of the bookstores.
GPT-4o also misclassified error types and failed to suggest refinements, as seen in the incorrectly tagged span ``\verb|<relation>street</relation>|''.
In contrast, \textsc{Zina} correctly detected the hallucinated spans and successfully provided appropriate refinements.

{In the example on the right, the original description contained hallucinations of the \facttype, \relationtype, and \numbertype types. 
GPT-4o failed to preserve the original phrasing—for example, modifying ``ticket vending machine'' to ``ticket machine'', which can be problematic in practical applications. In contrast, our two-step strategy delegates token copying to a deterministic function, which guarantees correctness by construction, as shown in Fig.~\ref{fig:results}. This property offers a clear advantage in real-world scenarios where it is crucial to edit only the erroneous parts while preserving the rest of the text exactly as is.}

\subsection{Ablation Studies}
\label{sec:ablation_studies}

Table~\ref{table:ablation} presents the quantitative results of the ablation studies.
To assess the contribution of each module in our proposed method, we conducted three ablation studies.

\vspace{-3mm}
\paragraph{Decoupling strategy ablation.}
To evaluate the contribution of our core strategy (i.e. decoupling detection and tagging), we replaced our pipeline with a single MLLM that performs both jointly.  
For fairness, Model(i) was fine-tuned with the same training budget as \textsc{Zina}.
As shown in Table~\ref{table:ablation}, the comparison between Model~(i) and Model~(vi) reveals a substantial performance drop, with the F$_1$ score for the detection task decreasing by 23.24 points, {partially due to exposure bias.}
Similarly, in the editing task, performance decreased by 28.48, 32.51, 2.76, and 2.67 points on BERT-F$_1$, CLIP-F$_1$, CLIP-S, and PAC-S, respectively.
These results highlight the effectiveness of our decoupling strategy in improving both detection and editing performance.

{Moreover, taken together with Table~\ref{table:quant}, these results indicate that the performance improvements mainly stem from the combination of synthetic data and the two-step generation strategy rather than either component in isolation.
One-step systems (e.g.~\cite{fava}) insert tags directly into the sentence; even a single misplaced tag can cause distributional shifts due to the autoregressive nature of LLMs.
In contrast, as discussed in Section~\ref{sec:methodlogy}, the two-step strategy mitigates exposure bias and enables more effective use of the synthetic dataset, as evidenced by \textsc{Zina} (Model~(vi)).}

\vspace{-3mm}
\paragraph{Backbone ablation.}
We investigated the effect of different backbones by replacing the Qwen2.5-VL-72B backbone with Qwen2.5-VL-32B and LLaVA-OV-72B \cite{llavaov}.
The latter was selected due to its strong performance and a model size comparable to Qwen2.5-VL-72B.
Table~\ref{table:ablation} shows that Model~(vi) outperformed both Model~(ii), which uses the Qwen2.5-VL-32B backbone, and Model~(iii), which uses the LLaVA-OV-72B backbone.
Specifically, Model~(vi) improved the F$_1$ score on the detection task by 12.60 points compared to Model~(ii).
Similarly, in the editing task, it also outperformed Model~(ii) by 16.5, 15.73, 4.88, and 1.73 points on BERT-F$_1$, CLIP-F$_1$, CLIP-S, and PAC-S, respectively.
These results demonstrate that the Qwen2.5-VL-72B backbone contributed to overall performance.

\vspace{-3mm}
\paragraph{Few-shot ablation.}
Table~\ref{table:ablation} shows that Model~(vi) outperformed Model~(iv) with $n = 1$ and Model~(v) with $n = 2$.  
Specifically, Model~(vi) achieved improvements of 0.94, 0.63, 0.18, 0.09, and 0.04 in F$_1$, BERT-F$_1$, CLIP-F$_1$, CLIP-S, and PAC-S, respectively, compared to Model~(v) with $n = 2$.
These results indicate that the current few-shot setting ($n = 3$) had a substantial impact on overall performance.

\vspace{-5mm}
\section{Conclusion}

We focused on the automatic fine-grained hallucination detection and editing for MLLMs.
The contributions of this paper are as follows:
(i) We proposed \textsc{Zina}, a novel method for fine-grained hallucination detection and editing.
(ii) We constructed the VisionHall dataset, which comprises outputs generated by twelve MLLMs, with hallucinated spans manually annotated according to our taxonomy.
(iii) \textsc{Zina} outperformed existing methods on VisionHall and MHaluBench.

\vspace{-3mm}
\paragraph{Limitations.}
While our method achieved strong performance on both detection and editing tasks, it has several limitations.
First, the two-stage architecture results in relatively high inference time, which may hinder deployment in real-time or resource-constrained settings.
Second, despite outperforming baseline models overall, our method often underdetects hallucinations. This may stem from limitations in the reviewer MLLM, which tends to be overly conservative in generating tags.

\vspace{-2mm}
\section*{Acknowledgements}
This work was partially supported by JSPS KAKENHI
Grant Number 23K28168, JST Moonshot, and JSPS
Fellows Grant Number JP25KJ2069.

{\small
\bibliographystyle{ieee_fullname}
\bibliography{reference}

@preamble{ "\providecommand{\CNFX}[1]{{\em{\textrm{(#1)}}}}"
}

@article{llama,
  author =        {Grattafiori, Aaron and Dubey, Abhimanyu and
                   Jauhri, Abhinav and Pandey, Abhinav and
                   Kadian, Abhishek and Al-Dahle, Ahmad and
                   Letman, Aiesha and Mathur, Akhil and Schelten, Alan and
                   Vaughan, Alex and others},
  journal =       {arXiv preprint arXiv:2407.21783},
  pages =         {arXiv--2407},
  title =         {{The llama 3 herd of models}},
  year =          {2024},
}

@inproceedings{llava-1.5,
  author =        {Liu, Haotian and Li, Chunyuan and Li, Yuheng and
                   Lee, Yong Jae},
  booktitle =     {CVPR},
  pages =         {26296-26306},
  title =         {{Improved Baselines with Visual Instruction Tuning}},
  year =          {2024},
}

@inproceedings{llava,
  author =        {Liu, Haotian and Li, Chunyuan and Wu, Qingyang and
                   Lee, Yong Jae},
  booktitle =     {NeurIPS},
  pages =         {34892--34916},
  title =         {{Visual Instruction Tuning}},
  year =          {2023},
}

@article{gpt4,
  author =        {Josh Achiam and Steven Adler and others},
  journal =       {arXiv preprint arXiv:2303.08774},
  title =         {{GPT-4 Technical Report}},
  year =          {2023},
}

@article{gemini,
  author =        {Gemini Team and Anil, Rohan and Borgeaud, Sebastian and
                   Alayrac, Jean-Baptiste and Yu, Jiahui and
                   Soricut, Radu and Schalkwyk, Johan and Dai, Andrew M. and
                   Hauth, Anja and Millican, Katie and Silver, David and
                   Johnson, Melvin and Antonoglou, Ioannis and
                   Schrittwieser, Julian and Glaese, Amelia},
  journal =       {arXiv preprint arXiv:2312.11805},
  title =         {{Gemini: A Family of Highly Capable Multimodal
                   Models}},
  year =          {2023},
}

@inproceedings{vila,
  author =        {Lin, Ji and Yin, Hongxu and Ping, Wei and
                   Molchanov, Pavlo and Shoeybi, Mohammad and Han, Song},
  booktitle =     {CVPR},
  pages =         {26679--26689},
  title =         {{VILA: On Pre-training for Visual Language Models}},
  year =          {2024},
}

@inproceedings{qwen-vl,
  author =        {Bai, Jinze and Bai, Shuai and Yang, Shusheng and
                   Wang, Shijie and Tan, Sinan and Wang, Peng and
                   Lin, Junyang and Zhou, Chang and Zhou, Jingren},
  booktitle =     {ICLR},
  title =         {{Qwen-VL: A Frontier Large Vision-Language Model with
                   Versatile Abilities}},
  year =          {2024},
}

@inproceedings{instructblip,
  author =        {Dai, Wenliang and Li, Junnan and Li, Dongxu and
                   Tiong, Anthony Meng Huat and Zhao, Junqi and
                   Wang, Weisheng and Li, Boyang and Fung, Pascale N and
                   Hoi, Steven},
  booktitle =     {NeurIPS},
  title =         {{InstructBLIP: Towards General-Purpose
                   Vision-Language Models with Instruction Tuning}},
  year =          {2023},
}

@article{amber,
  author =        {Wang, Junyang and Wang, Yuhang and Xu, Guohai and
                   Zhang, Jing and Gu, Yukai and Jia, Haitao and
                   Wang, Jiaqi and Xu, Haiyang and Yan, Ming and
                   Zhang, Ji and others},
  journal =       {arXiv preprint arXiv:2311.07397},
  title =         {{Amber: An llm-free multi-dimensional benchmark for
                   mllms hallucination evaluation}},
  year =          {2023},
}

@inproceedings{mhaldetect,
  author =        {Gunjal, Anisha and Yin, Jihan and Bas, Erhan},
  booktitle =     {AAAI},
  pages =         {18135--18143},
  title =         {{Detecting and preventing hallucinations in large
                   vision language models}},
  year =          {2024},
}

@inproceedings{unihd,
  address =       {Bangkok, Thailand},
  author =        {Chen, Xiang and Wang, Chenxi and Xue, Yida and
                   Zhang, Ningyu and Yang, Xiaoyan and Li, Qiang and
                   Shen, Yue and Liang, Lei and Gu, Jinjie and
                   Chen, Huajun},
  booktitle =     {Proceedings of the 62nd Annual Meeting of the
                   Association for Computational Linguistics (Volume 1:
                   Long Papers)},
  editor =        {Ku, Lun-Wei and Martins, Andre and Srikumar, Vivek},
  month =         aug,
  pages =         {3235--3252},
  publisher =     {Association for Computational Linguistics},
  title =         {{Unified Hallucination Detection for Multimodal Large
                   Language Models}},
  year =          {2024},
  doi =           {10.18653/v1/2024.acl-long.178},
  url =           {https://aclanthology.org/2024.acl-long.178/},
}

@inproceedings{pope,
  author =        {Li, Yifan and Du, Yifan and Zhou, Kun and
                   Wang, Jinpeng and Zhao, Xin and Wen, Ji-Rong},
  booktitle =     {EMNLP},
  pages =         {292--305},
  title =         {{Evaluating Object Hallucination in Large
                   Vision-Language Models}},
  year =          {2023},
}

@inproceedings{spice,
  author =        {Anderson, Peter and Fernando, Basura and
                   Johnson, Mark and Gould, Stephen},
  booktitle =     {ECCV},
  pages =         {382--398},
  title =         {{SPICE: Semantic propositional image caption
                   evaluation}},
  year =          {2016},
}

@inproceedings{fava,
  author =        {Mishra, Abhika and Asai, Akari and
                   Balachandran, Vidhisha and Wang, Yizhong and
                   Neubig, Graham and Tsvetkov, Yulia and
                   Hajishirzi, Hannaneh},
  booktitle =     {COLM},
  title =         {{Fine-grained hallucination detection and editing for
                   language models}},
  year =          {2024},
}

@inproceedings{HalluMeasure,
  author =        {Akbar, Shayan Ali and Hossain, Md Mosharaf and
                   Wood, Tess and Chin, Si-Chi and Salinas, Erica M and
                   Alvarez, Victor and Cornejo, Erwin},
  booktitle =     {EMNLP},
  pages =         {15020--15037},
  title =         {{H}allu{M}easure: Fine-grained Hallucination
                   Measurement Using Chain-of-Thought Reasoning},
  year =          {2024},
  doi =           {10.18653/v1/2024.emnlp-main.837},
  url =           {https://aclanthology.org/2024.emnlp-main.837/},
}

@article{manakul2023selfcheckgpt,
  author =        {Manakul, Potsawee and Liusie, Adian and
                   Gales, Mark JF},
  journal =       {Proceedings of the 2023 Conference on Empirical
                   Methods in Natural Language Processing},
  title =         {Selfcheckgpt: Zero-resource black-box hallucination
                   detection for generative large language models},
  year =          {2023},
  url =           {https://aclanthology.org/2023.emnlp-main.557.pdf},
}

@inproceedings{min2023factscore,
  author =        {Min, Sewon and Krishna, Kalpesh and Lyu, Xinxi and
                   Lewis, Mike and Yih, Wen-tau and Koh, Pang Wei and
                   Iyyer, Mohit and Zettlemoyer, Luke and
                   Hajishirzi, Hannaneh},
  booktitle =     {EMNLP},
  title =         {{FActScore}: Fine-grained Atomic Evaluation of
                   Factual Precision in Long Form Text Generation},
  year =          {2023},
  url =           {https://aclanthology.org/2023.emnlp-main.741/},
}

@inproceedings{gao2022rarr,
  author =        {Gao, Luyu and Dai, Zhuyun and Pasupat, Panupong and
                   Chen, Anthony and Chaganty, Arun Tejasvi and
                   Fan, Yicheng and Zhao, Vincent Y and Lao, Ni and
                   Lee, Hongrae and Juan, Da-Cheng and others},
  booktitle =     {Proceedings of the 61st Annual Meeting of the
                   Association for Computational Linguistics},
  title =         {{RARR}: Researching and revising what language models
                   say, using language models},
  year =          {2022},
  url =           {https://aclanthology.org/2023.acl-long.910/},
}

@article{chen2023purr,
  author =        {Chen, Anthony and Pasupat, Panupong and Singh, Sameer and
                   Lee, Hongrae and Guu, Kelvin},
  journal =       {arXiv preprint arXiv:2305.14908},
  title =         {PURR: Efficiently Editing Language Model
                   Hallucinations by Denoising Language Model
                   Corruptions},
  year =          {2023},
  url =           {https://arxiv.org/abs/2305.14908},
}

@article{li2024dawndarkempiricalstudy,
  author =        {Li, Junyi and Chen, Jie and Ren, Ruiyang and
                   Cheng, Xiaoxue and Zhao, Wayne Xin and Nie, Jian-Yun and
                   Wen, Ji-Rong},
  journal =       {arXiv preprint arXiv:2401.03205},
  title =         {The dawn after the dark: An empirical study on
                   factuality hallucination in large language models},
  year =          {2024},
}

@article{cceval,
  author =        {Zhai, Bohan and Yang, Shijia and Xu, Chenfeng and
                   Shen, Sheng and Keutzer, Kurt and Li, Chunyuan and
                   Li, Manling},
  journal =       {arXiv preprint arXiv:2310.01779},
  title =         {{HallE-Control: controlling object hallucination in
                   large multimodal models}},
  year =          {2023},
}

@article{chair,
  author =        {Rohrbach, Anna and Hendricks, Lisa Anne and
                   Burns, Kaylee and Darrell, Trevor and Saenko, Kate},
  journal =       {arXiv preprint arXiv:1809.02156},
  title =         {{Object hallucination in image captioning}},
  year =          {2018},
}

@inproceedings{halloc,
  author =        {Park, Eunkyu and Kim, Minyeong and Kim, Gunhee},
  booktitle =     {CVPR},
  pages =         {29893--29903},
  title =         {{Halloc: Token-level localization of hallucinations
                   for vision language models}},
  year =          {2025},
}

@inproceedings{esreal,
  author =        {Kim, Minchan and Kim, Minyeong and Bae, Junik and
                   Choi, Suhwan and Kim, Sungkyung and Chang, Buru},
  booktitle =     {ECCV},
  pages =         {236--252},
  title =         {{Exploiting semantic reconstruction to mitigate
                   hallucinations in vision-language models}},
  year =          {2024},
}

@inproceedings{dci,
  author =        {Urbanek, Jack and Bordes, Florian and Astolfi, Pietro and
                   Williamson, Mary and Sharma, Vasu and
                   Romero-Soriano, Adriana},
  booktitle =     {CVPR},
  pages =         {26700-26709},
  title =         {{A Picture is Worth More Than 77 Text Tokens:
                   Evaluating CLIP-Style Models on Dense Captions}},
  year =          {2024},
}

@article{qwen25vl,
  author =        {Bai, Shuai and Chen, Keqin and Liu, Xuejing and
                   Wang, Jialin and Ge, Wenbin and Song, Sibo and
                   Dang, Kai and Wang, Peng and Wang, Shijie and
                   Tang, Jun and Zhong, Humen and Zhu, Yuanzhi and
                   Yang, Mingkun and Li, Zhaohai and Wan, Jianqiang and
                   Wang, Pengfei and Ding, Wei and Fu, Zheren and
                   Xu, Yiheng and Ye, Jiabo and Zhang, Xi and
                   Xie, Tianbao and Cheng, Zesen and Zhang, Hang and
                   Yang, Zhibo and Xu, Haiyang and Lin, Junyang},
  journal =       {arXiv preprint arXiv:2502.13923},
  title =         {Qwen2.5-VL Technical Report},
  year =          {2025},
}

@inproceedings{schuster-etal-2021-get2,
  author =        {Schuster, Tal and Fisch, Adam and Barzilay, Regina},
  booktitle =     {NAACL},
  title =         {Get Your Vitamin {C}! Robust Fact Verification with
                   Contrastive Evidence},
  year =          {2021},
}

@inproceedings{feng2023factkb,
  author =        {Feng, Shangbin and Balachandran, Vidhisha and
                   Bai, Yuyang and Tsvetkov, Yulia},
  booktitle =     {EMNLP},
  title =         {{F}act{KB}: Generalizable Factuality Evaluation using
                   Language Models Enhanced with Factual Knowledge},
  year =          {2023},
}

@inproceedings{clipscore,
  author =        {Hessel, Jack and Holtzman, Ari and Forbes, Maxwell and
                   Le Bras, Ronan and Choi, Yejin},
  booktitle =     {EMNLP},
  pages =         {7514--7528},
  title =         {{CLIPScore: A Reference-free Evaluation Metric for
                   Image Captioning}},
  year =          {2021},
}

@inproceedings{pacs,
  author =        {Sarto, Sara and Barraco, Manuele and Cornia, Marcella and
                   Baraldi, Lorenzo and Cucchiara, Rita},
  booktitle =     {CVPR},
  pages =         {6914--6924},
  title =         {{Positive-Augmented Contrastive Learning for Image
                   and Video Captioning Evaluation}},
  year =          {2023},
}

@inproceedings{polos,
  author =        {Wada, Yuiga and Kanta, Kaneda and Daichi, Saito and
                   Komei Sugiura},
  booktitle =     {CVPR},
  pages =         {13559--13568},
  title =         {{Polos: Multimodal Metric Learning from Human
                   Feedback for Image Captioning}},
  year =          {2024},
}

@inproceedings{vela,
  author =        {Matsuda, Kazuki and others},
  booktitle =     {EMNLP},
  pages =         {8691--8707},
  title =         {{VELA: An LLM-Hybrid-as-a-Judge Approach for
                   Evaluating Long Image Captions}},
  year =          {2025},
}

@inproceedings{pearl,
  author =        {Hirano, Shinnosuke and Wada, Yuiga and
                   Matsuda, Kazuki and Otsuki, Seitaro and
                   Sugiura, Komei},
  booktitle =      {AAAI},
  title =         {{LLM-Free Image Captioning Evaluation in
                   Reference-Flexible Settings}},
  year =          {2026},
}

@inproceedings{bert,
  author =        {Devlin, Jacob and Chang, Ming-Wei and others},
  booktitle =     {NAACL},
  pages =         {4171--4186},
  title =         {{BERT: Pre-training of Deep Bidirectional
                   Transformers for Language Understanding}},
  year =          {2019},
}

@inproceedings{clip,
  author =        {Radford, Alec and Kim, Jong Wook and Hallacy, Chris and
                   Ramesh, Aditya and Goh, Gabriel and Agarwal, Sandhini and
                   Sastry, Girish and Askell, Amanda and Mishkin, Pamela and
                   Clark, Jack and Krueger, Gretchen and
                   Sutskever, Ilya},
  booktitle =     {ICML},
  pages =         {8748--8763},
  title =         {{Learning Transferable Visual Models from Natural
                   Language Supervision}},
  year =          {2021},
}

@inproceedings{bertscore,
  author =        {Zhang, Tianyi and Kishore, Varsha and Wu, Felix and
                   Weinberger, Kilian and Artzi, Yoav},
  booktitle =     {ICLR},
  title =         {{BERTScore: Evaluating Text Generation with BERT}},
  year =          {2020},
}

@article{asai2023self,
  author =        {Asai, Akari and Wu, Zeqiu and Wang, Yizhong and
                   Sil, Avirup and Hajishirzi, Hannaneh},
  journal =       {arXiv preprint arXiv:2310.11511},
  title =         {{Self-RAG}: Learning to Retrieve, Generate, and
                   Critique through Self-Reflection},
  year =          {2023},
  url =           {https://arxiv.org/abs/2310.11511},
}

@inproceedings{coco,
  author =        {Lin, Tsung and Maire, Michael and Belongie, Serge and
                   Bourdev, Lubomir and Girshick, Ross and Hays, James and
                   Perona, Pietro and Ramanan, Deva and
                   Zitnick, C. Lawrence and Dollar, Piotr},
  booktitle =     {ECCV},
  pages =         {740--755},
  title =         {{Microsoft {COCO}: Common Objects in Context}},
  year =          {2014},
}

@inproceedings{nocaps,
  author =        {Agrawal, Harsh and Desai, Karan and Wang, Yufei and
                   Chen, Xinlei and Jain, Rishabh and Johnson, Mark and
                   Batra, Dhruv and Parikh, Devi and Lee, Stefan and
                   Anderson, Peter},
  booktitle =     {ICCV},
  pages =         {8948--8957},
  title =         {{nocaps: Novel Object Captioning at Scale}},
  year =          {2019},
}

@misc{o3,
  author =        {{OpenAI}},
  month =         {April},
  title =         {{OpenAI o3 and o4-mini System Card}},
  year =          {2025},
  url =           {https://openai.com/index/o3-o4-mini-system-card/},
}

@inproceedings{zhong-etal-2024-investigating,
  author =        {Zhong, Weihong and Feng, Xiaocheng and Zhao, Liang and
                   Li, Qiming and Huang, Lei and Gu, Yuxuan and
                   Ma, Weitao and Xu, Yuan and Qin, Bing},
  booktitle =     {ACL},
  pages =         {11991--12011},
  title =         {{Investigating and Mitigating the Multimodal
                   Hallucination Snowballing in Large Vision-Language
                   Models}},
  year =          {2024},
}

@article{qwen2vl,
  author =        {Wang, Peng and Bai, Shuai and Tan, Sinan and
                   Wang, Shijie and Fan, Zhihao and Bai, Jinze and
                   Chen, Keqin and Liu, Xuejing and Wang, Jialin and
                   Ge, Wenbin and others},
  journal =       {arXiv preprint arXiv:2409.12191},
  title =         {{Qwen2-VL: Enhancing vision-language model's
                   perception of the world at any resolution}},
  year =          {2024},
}

@article{llavaov,
  author =        {Li, Bo and Zhang, Yuanhan and Guo, Dong and
                   Zhang, Renrui and Li, Feng and Zhang, Hao and
                   Zhang, Kaichen and Li, Yanwei and Liu, Ziwei and
                   Li, Chunyuan},
  journal =       {arXiv preprint arXiv:2408.03326},
  title =         {{LLaVA-OneVision: Easy Visual Task Transfer}},
  year =          {2024},
}

@misc{llavanext,
  author =        {Liu, Haotian and Li, Chunyuan and Li, Yuheng and
                   Li, Bo and Zhang, Yuanhan and Shen, Sheng and
                   Lee, Jae},
  title =         {{LLaVA-NeXT: Improved reasoning, OCR, and world
                   knowledge}},
  year =          {2024},
}

@inproceedings{bartscore,
  author =        {Yuan, Weizhe and Neubig, Graham and others},
  booktitle =     {NeurIPS},
  pages =         {27263--27277},
  title =         {{BARTScore: Evaluating Generated Text as Text
                   Generation}},
  volume =        {34},
  year =          {2021},
}

@article{mmgpt,
  author =        {Gong, Tao and Lyu, Chengqi and Zhang, Shilong and
                   Wang, Yudong and Zheng, Miao and Zhao, Qian and
                   Liu, Kuikun and Zhang, Wenwei and Luo, Ping and
                   Chen, Kai},
  journal =       {arXiv preprint arXiv:2305.04790},
  title =         {{MultiModal-GPT: A Vision and Language Model for
                   Dialogue with Humans}},
  year =          {2023},
}

@inproceedings{sharegpt4v,
  author =        {Chen, Lin and Li, Jinsong and Dong, Xiaoyi and
                   Zhang, Pan and He, Conghui and Wang, Jiaqi and
                   Zhao, Feng and Lin, Dahua},
  booktitle =     {ECCV},
  pages =         {370--387},
  title =         {{ShareGPT4V: Improving Large Multi-Modal Models with
                   Better Captions}},
  year =          {2024},
}

@inproceedings{internvl,
  author =        {Chen, Zhe and Wu, Jiannan and Wang, Wenhai and
                   Su, Weijie and Chen, Guo and Xing, Sen and
                   Zhong, Muyan and Zhang, Qinglong and Zhu, Xizhou and
                   Lu, Lewei and Li, Bin and Luo, Ping and Lu, Tong and
                   Qiao, Yu and Dai, Jifeng},
  booktitle =     {CVPR},
  pages =         {24185--24198},
  title =         {{InternVL: Scaling up Vision Foundation Models and
                   Aligning for Generic Visual-Linguistic Tasks}},
  year =          {2024},
}

@inproceedings{blip2,
  author =        {Li, Junnan and Li, Dongxu and Savarese, Silvio and
                   Hoi, Steven},
  booktitle =     {ICML},
  pages =         {19730--19742},
  title =         {{BLIP-2: Bootstrapping Language-Image Pre-training
                   with Frozen Image Encoders and Large Language
                   Models}},
  year =          {2023},
}

@article{git,
  author =        {Jianfeng Wang and Zhengyuan Yang and Xiaowei Hu and
                   Linjie Li and Kevin Lin and Zhe Gan and Zicheng Liu and
                   Ce Liu and Lijuan Wang},
  journal =       {TMLR},
  note =          {},
  title =         {{{GIT}: A Generative Image-to-text Transformer for
                   Vision and Language}},
  year =          {2022},
  issn =          {2835-8856},
  url =           {https://openreview.net/forum?id=b4tMhpN0JC},
}

@article{deberta,
  author =        {He, Pengcheng and Liu, Xiaodong and Gao, Jianfeng and
                   Chen, Weizhu},
  journal =       {arXiv preprint arXiv:2006.03654},
  title =         {{DeBERTa: Decoding-enhanced bert with disentangled
                   attention}},
  year =          {2020},
}

@inproceedings{foil,
  author =        {Shekhar, Ravi and Pezzelle, Sandro and
                   Klimovich, Yauhen and others},
  booktitle =     {ACL},
  pages =         {255--265},
  title =         {{{FOIL} it! Find One Mismatch Between Image and
                   Language caption}},
  year =          {2017},
}

@inproceedings{yao2024hifi,
  author =        {Yao, Ziwei and Wang, Ruiping and Chen, Xilin},
  booktitle =     {ECCV},
  pages =         {441--458},
  title =         {{HiFi-Score: Fine-Grained Image Description
                   Evaluation with Hierarchical Parsing Graphs}},
  year =          {2024},
}

@inproceedings{med,
  author =        {Tianyi Bai and Yuxuan Fan and Qiu Jiantao and others},
  booktitle =     {NeurIPS},
  title =         {{Hallucination at a Glance: Controlled Visual Edits
                   and Fine-Grained Multimodal Learning}},
  year =          {2025},
}

@inproceedings{lora,
  author =        {Hu, Edward J and Shen, Yelong and Wallis, Phillip and
                   Allen-Zhu, Zeyuan and Li, Yuanzhi and Wang, Shean and
                   Wang, Lu and Chen, Weizhu and others},
  booktitle =     {ICLR},
  title =         {{LoRA: Low-rank Adaptation of Large Language Models}},
  year =          {2022},
}

@inproceedings{llama-factory,
  address =       {Bangkok, Thailand},
  author =        {Yaowei Zheng and Richong Zhang and Junhao Zhang and
                   Yanhan Ye and Zheyan Luo and Zhangchi Feng and
                   Yongqiang Ma},
  booktitle =     {ACL (Volume 3: System Demonstrations)},
  publisher =     {ACL},
  title =         {{LlamaFactory: Unified Efficient Fine-Tuning of 100+
                   Language Models}},
  year =          {2024},
}

@inproceedings{vlmevalkit,
  author =        {Duan, Haodong and Yang, Junming and Qiao, Yuxuan and
                   Fang, Xinyu and Chen, Lin and Liu, Yuan and
                   Dong, Xiaoyi and Zang, Yuhang and Zhang, Pan and
                   Wang, Jiaqi and others},
  booktitle =     {Proceedings of the 32nd ACM International Conference
                   on Multimedia},
  pages =         {11198--11201},
  title =         {Vlmevalkit: An open-source toolkit for evaluating
                   large multi-modality models},
  year =          {2024},
}

@inproceedings{long-clip,
  author =        {Zhang, Beichen and Zhang, Pan and Dong, Xiaoyi and
                   Zang, Yuhang and Wang, Jiaqi},
  booktitle =     {ECCV},
  pages =         {310--325},
  title =         {{Long-CLIP: Unlocking the Long-Text Capability of
                   CLIP}},
  year =          {2024},
}
}

\clearpage
\appendix
\setcounter{page}{1}
\maketitlesupplementary

\section{Taxonomy Construction}
\label{sec:taxonomy_construction}

{{We designed the taxonomy in three steps:}
{\paragraph{1. Initial Design.} Since our focus is on hallucinations in image captioning, we first drew inspiration from standard image captioning evaluation, specifically SPICE~\cite{spice}. SPICE defines object, attribute, and relation as the core components of image captions. Given their established role in image captioning, we adopted these three as candidate categories. We further referred to existing taxonomies such as POPE~\cite{pope}, AMBER~\cite{amber}, and UniHD~\cite{unihd}. While POPE and AMBER include object and attribute, UniHD additionally introduces text and fact hallucinations. Based on these works, we initially selected five categories.}
{\paragraph{2. Pilot Annotation Experiment.} To check the taxonomy's coverage, we then conducted a pilot annotation experiment of 200 samples by a small group of annotators. Feedback from this experiment revealed frequent errors related to number and grammar. Since grammatical errors are generally excluded from hallucination definitions, we added number as a new candidate.}
{\paragraph{3. Validation through Large-scale Annotation.} To validate the final taxonomy, we performed a crowdsourced annotation task for about 2k samples. As shown in Table~\ref{table:taxonomy}, even GPT-4o exhibited hallucinations across all six categories, supporting the validity of our taxonomy.}

\section{Evaluation Metrics}
\label{appendix:metrics}

\paragraph{Span-level metrics.}
By utilizing \cite{bertscore, clipscore}, BERT-F$_1$ and CLIP-F$_1$ are computed by setting the similarity function $\mathrm{sim}(\cdot,\cdot)$ to $\mathrm{BERTScore}(\cdot,\cdot)$ and $\mathrm{CLIPScore}(\cdot,\cdot)$, respectively. The precision, recall, and F$_1$ score are defined as follows:
\begin{align}
&\mathrm{Precision} = \frac{\sum_{i \in N} \mathrm{sim}(\hat{y}_{\mathrm{text}}^{(i)}, y_{\mathrm{text}}^{(i)})}{\sum_{i \in N} \mathbf{1}[\hat{y}_{\mathrm{type}}^{(i)} \ne 0]}, \\
&\mathrm{Recall} = \frac{\sum_{i \in N} \mathrm{sim}(\hat{y}_{\mathrm{text}}^{(i)}, y_{\mathrm{text}}^{(i)})}{\sum_{i \in N} \mathbf{1}[y_{\mathrm{type}}^{(i)} \ne 0]}, \\
&\mathrm{F}_1 = 2 \cdot \frac{\mathrm{Precision} \cdot \mathrm{Recall}}{\mathrm{Precision} + \mathrm{Recall}}.
\end{align}
Following \cite{bertscore}, we used DeBERTa embeddings \cite{deberta} for computing BERTScore\footnote{\url{https://huggingface.co/microsoft/deberta-xlarge-mnli}}.

\paragraph{Sentence-level metrics.}
We employed CLIP-S~\cite{clipscore} and PAC-S~\cite{pacs} for sentence-level evaluation.  We selected them because they are standard metrics for image captioning evaluation.

{
CLIP-S first obtains the embeddings $t$, $v$, and $r$ from CLIP \cite{clip} for the edited text, image, and reference, respectively, and then computes the mean of the cosine similarities.
The score $\mathrm{Score}(t,v,r)$ of CLIP-S is defined as follows:}
\begin{align}
S(x,y) = w \cdot \max\!\big(\cos(x,y),\,0\big), \\
\mathrm{Score}(t,v,r) = \mathrm{mean}\big(S(t,v),\, S(t,r)\big),
\end{align}
{
where $w$ denotes a rescaling factor of 2.5~\cite{clipscore}.
Similarly, PAC-S adopts the same form but uses a CLIP fine-tuned for the task-specific domain.
These metrics assign a single score to the edited text, indicating how well it aligns with both the image and the reference.
}

{
As sentence-level metrics are standard in hallucination editing evaluation~\cite{fava, foil}, we report CLIP-S and PAC-S alongside BERT-F$_1$ and CLIP-F$_1$.
CLIP-S and PAC-S rely on CLIP~\cite{clip}, which limits input text to a maximum of 77 tokens.
Therefore, previous works \cite{yao2024hifi, vela} used modified variants of these metrics. These variants compute the cosine similarity between each sentence in a paragraph and the image, and then average the values to obtain the final score.
Following this approach, we applied the same sentence-wise averaging for CLIP-S and PAC-S.
}

\begin{figure}[t]
    \centering      \includegraphics[width=1.0\linewidth]{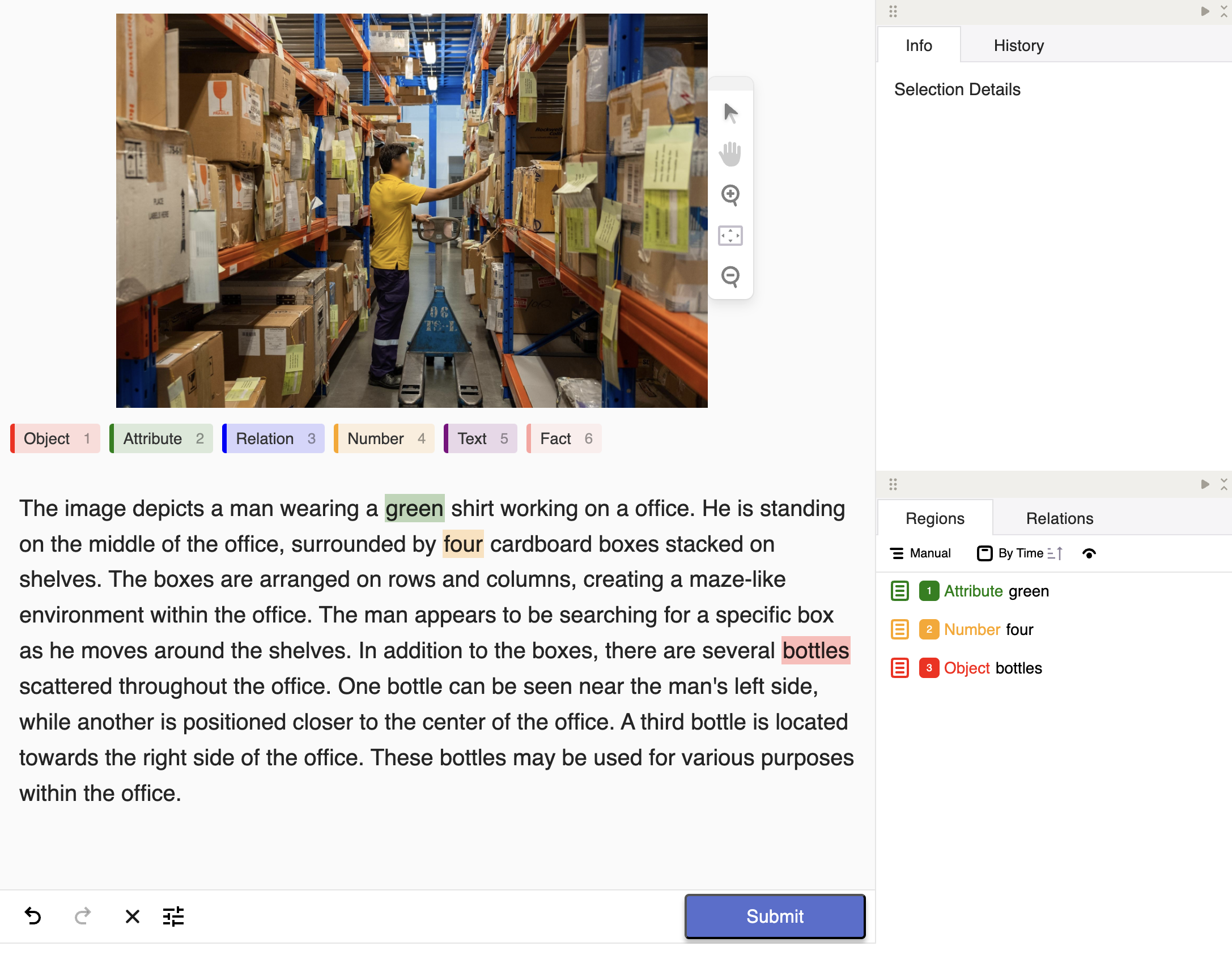}
    \caption{Annotation interface used for the VisionHall dataset.}
    \label{fig:interface}
\end{figure}

\section{Construction of VisionHall Dataset}
\label{sec:construction}

\paragraph{Text-editing vs. image-editing.}
Although modifying text to inject hallucinations is the standard approach~\cite{fava,halloc}, some studies (e.g. \cite{med}) generate hallucinations via image editing. These image-editing approaches are problematic because they can introduce image-text discrepancies through uncontrolled visual artifacts (e.g. alterations in non-target regions). In contrast, text-editing methods modify textual descriptions, enabling better-controlled errors; we therefore follow this text-editing approach.

\paragraph{Annotation Process.}
We employed LabelStudio \footnote{\url{https://github.com/HumanSignal/label-studio}} for collecting the VisionHall dataset.
Fig.~\ref{fig:interface} illustrates the annotation interface.
We conducted annotations via a public crowdsourcing platform, where we recruited annotators from a broad online population without imposing restrictions on demographic or geographic background.
We recruited annotators and provided payment that was adequate based on the participants' country of residence.
We also obtained consent through the task instructions, which explicitly stated that the collected data would be used for research purposes.

The full text of the instructions given to participants is as follows:
\TColorBox{Annotation Instructions}{
You will be given an image along with a caption describing its content. Your task is to identify any parts of the caption that contain hallucinations and label them using one of the six predefined categories:
 \\
 \\- Spatial Relation Hallucination
 \\    - Errors in spatial relationships  (Note: Grammatical mistakes in prepositions do NOT count as hallucinations.)
 \\    - For example:
 \\    If the caption says "an apple on the table", but the apple is not actually on the table, you should select the word "on".
 \\    If the caption says "an apple in to the table", the phrase "in to" is a grammatical error, so you should not select any words.
 \\
 \\- Attribute Hallucination <attribute>
 \\    - Errors in descriptive attributes, such as adjectives or adverbs.
 \\    - For example:
 \\    If the caption says "a blue apple", but the apple is not actually blue, you should select the word "blue".
 \\
 \\- Object Hallucination <object>
 \\    - Errors in naming specific objects or entities.
 \\    - For example:
 \\    If the caption says "a red apple", but there are no apples and instead a book is present, you should select the word "apple".
 \\- Number Hallucination <number>
 \\    - Errors related to numerical values or quantities.
 \\    - For example:
 \\    If the caption says "two red apples", but there are actually three apples, you should select the word "two".
 \\    
 \\- Factual Hallucination <fact>
 \\    - Errors concerning named entities (e.g., incorrect facts or identities).
 \\    - For example:
 \\    If the caption says "Steve Jobs with his family", but Steve Jobs is not present and instead it's Bill Gates, you should select "Steve Jobs".
 \\
 \\- Text Hallucination <text>
 \\    - Errors involving scene text (i.e., text appearing within the image).
 \\    - For example:
 \\    If the caption says "a sign that says STOP", but the sign actually says "GO", you should select the word "STOP".
 \\
 \\To label a word as a hallucination, simply click on it. To change the category of the tag, click on the "Categories" section shown in the interface. You can ignore grammatical mistakes - they are not considered hallucinations.
}

\begin{table}[t]
  \centering
  \begin{tabular}{lc}
    \hline
    Metric & Value \\
    \hline
    Total graphs                & 108,918 \\
    Total nodes                 & 130,968 \\
    Avg.\ connected components  & 5.45 \\
    Avg.\ nodes                 & 6.55 \\
    Avg.\ edges                 & 2.53 \\
    Avg.\ degree                & 2.06 \\
    \hline
  \end{tabular}
    \caption{Structural statistics of the pruned dependency graphs}
  \label{tab:vh_struct}
\end{table}
\begin{table*}[t]
  \centering
  \setlength{\tabcolsep}{6pt}
\begin{tabular}{lccccc}
    \toprule
    \textbf{Model} & \textbf{Detection F$_1$} & \textbf{BERT-F$_1$} & \textbf{CLIP-F$_1$} & \textbf{CLIP-S} & \textbf{PAC-S} \\
    \midrule
    (i) \(d < 2\)      & 37.92 & 34.48 & 40.70 & 65.53 & 74.09 \\
    (ii) \(p = 0.25\)   & 42.90 & 37.59 & 43.87 & 65.52 & 74.06 \\
    (iii) \textsc{Zina}  & \textbf{45.15} & \textbf{44.02} & \textbf{50.39} & \textbf{66.08} & \textbf{74.36} \\
    \bottomrule
  \end{tabular}
\caption{Ablation on graph-pruning strategies.
 (i) degree-based pruning with threshold \(d<2\) (from Table~\ref{tab:vh_struct});
 (ii) random pruning with probability \(p=0.25\);
 (iii) original \textsc{Zina}. \textsc{Zina} outperformed both pruning variants, while degree-based performs the worst.}
\label{tab:curation}
\end{table*}

\paragraph{Ethical considerations.}
The collected data contains no personally identifiable or offensive content. 
All data used in this study are publicly available. 
We also confirmed that the source websites, repositories, and publications include no statements indicating concerns about personal information.

\section{Details of Datasets}
\label{sec:statistics}

\subsection{VisionHall Dataset}
VisionHall comprises 6,854 MLLM-generated descriptions for 4,759 images, collected from 211 annotators, for the detection task.
For the editing task, VisionHall contains 20k synthetic samples by our novel graph-based method.
Each sample includes an English reference caption, with a total of 986,488 words, an average length of 143.9 words, and a vocabulary size of 21,757.
The MLLM-generated descriptions are also in English, totaling 1,189,673 words, with an average length of 173.6 words and a vocabulary size of 42,446. 
The test set for both tasks consists of 400 samples collected from the human annotators. Its size is comparable to that of MHaluBench~\cite{unihd}, whose test set for image captioning contains 200 samples. We used the training set to train the model and the test set to evaluate model performance.

{As shown in Table~\ref{tab:len_freq}, for most categories, the span lengths are generally comparable. Regarding tag frequency, we found that the overall distribution was also similar across real and synthetic data. One exception is the \attributetype category, which appears less frequently in the synthetic set. This is because the Error Insertion module is intentionally designed to frequently select rarer categories such as \facttype or \texttype, in order to improve coverage across diverse error types. As a result, attribute hallucinations are relatively underrepresented. These analyses suggest that the overall distribution gap remains relatively small in both span length and tag frequency.}

{Table~\ref{tab:vh_struct} presents structural statistics of the pruned dependency graphs, including the total number of graphs and nodes, as well as the average number of connected components, nodes, degree, and edges per graph. Each sample contains an average of 5.45 connected components, with each graph consisting of an average of 6.55 nodes and 2.53 edges. On average, each node is connected to 2.06 other nodes, suggesting sufficient structural complexity in the constructed graphs.}

\subsection{MHaluBench}
MHaluBench~\cite{unihd} is a benchmark for evaluating coarse-grained hallucination detection.
This dataset provides samples with coarse-grained annotations in both Image-to-Text and Text-to-Image settings.
We evaluated the baselines and our proposed method on the ``Image-to-Text'' subset of MHaluBench, as our task focuses on image captioning.
This subset includes 200 samples for the image captioning task and 200 samples for the VQA task.
The generated texts are collected using MLLMs such as LLaVA~\cite{llava}.

\paragraph{Evaluation.} MHaluBench provides both segment-level and claim-level annotations.
For the segment-level task, each sentence in the generated texts was labeled as either hallucinatory or non-hallucinatory.
For the claim-level task, claims extracted from these sentences were further annotated with hallucination labels.
An entire response is labeled hallucinatory if it contains at least one hallucinatory segment.

\paragraph{Baselines.} The authors of MHaluBench used two baselines: a Gemini-based Self-Check and a GPT-based Self-Check. Self-Check~\cite{unihd} detects hallucinations using chain-of-thought reasoning without relying on external knowledge. Following this setup, we employed these two baselines, as well as UniHD, on MHaluBench.

\section{Implementation Details}
\label{appendix:impl}

\begin{table*}[t]
	\centering
	\normalsize
\setlength{\tabcolsep}{6pt}
	\begin{tabular}{lcccccc}
		\toprule
        \multirow{2}{*}{\textbf{Method}}
		& \multirow{2}{*}{\textbf{Two-stage}}
		& \multicolumn{1}{c}{\textbf{Detection}}
		& \multicolumn{2}{c}{\textbf{Editing}}
		& \multicolumn{2}{c}{\textbf{Overall}} \\
		\cmidrule(lr){3-3}\cmidrule(lr){4-5}\cmidrule(lr){6-7}
		&
		& \textbf{F$_1$}
		& \textbf{CLIP-S} & \textbf{PAC-S}
		& \textbf{BERT-F$_1$} & \textbf{CLIP-F$_1$} \\
		\midrule

LLaVA-1.5-7B~\cite{llava-1.5} &  & 0.82 & 64.01 & 72.72 & 0.66  & 0.93   \\
		LLaVA-1.5-7B~\cite{llava-1.5} & $\checkmark$ & 3.39 & 64.41 & 72.70  & 3.39  & 3.39  \\

Qwen2-VL-7B~\cite{qwen2vl} &  & 3.36  & 64.79 & 73.01  & 3.62  & 4.98  \\
		Qwen2-VL-7B~\cite{qwen2vl} & $\checkmark$ & 13.13 & 64.70 & 73.04  & 13.79 & 16.41 \\

LLaVA-OV-Qwen2-7B~\cite{llavaov} & & 3.39  & 64.06 & 72.40 & 3.39  & 3.39   \\
		LLaVA-OV-Qwen2-7B~\cite{llavaov} & $\checkmark$ & 5.38  & 64.40 & 72.73 & 4.58  & 5.33  \\

LLaVA-1.5-13B~\cite{llava-1.5} &  & 4.73 & 64.74 & 73.02  & 5.08  & 6.71   \\
		LLaVA-1.5-13B~\cite{llava-1.5} & $\checkmark$ & 4.14 & 64.46 & 72.72 & 3.87  & 4.05   \\

LLaVA-NeXT-Qwen-32B~\cite{llavanext} &  & 19.09 & 65.34 & 73.47 & 24.29 & {31.06}  \\
		LLaVA-NeXT-Qwen-32B~\cite{llavanext} & $\checkmark$ & 12.75 & 64.40 & 72.69 & 9.49  & 12.60  \\

Llama-3.2-90B-Vision-Instruct~\cite{llama} &  & 16.92 & 65.28 & 73.54 & 14.56 & 17.62  \\
		Llama-3.2-90B-Vision-Instruct~\cite{llama} & $\checkmark$ & 16.74 & 64.93 & 73.40  & 15.30 & 17.20 \\

Qwen2.5-VL-72B-Instruct~\cite{qwen25vl} &  & 21.31 & 64.38 & 72.99 & 18.85 & 23.67  \\
		Qwen2.5-VL-72B-Instruct~\cite{qwen25vl} & $\checkmark$ & \underline{35.21} & 65.29 & 73.65 & \underline{34.06} & \underline{39.26}  \\

LLaVA-OV-Qwen2-72B~\cite{llavaov} &  & 25.70 & {65.74} & 73.91 & 20.81 & 26.81  \\
		LLaVA-OV-Qwen2-72B~\cite{llavaov} & $\checkmark$ & 34.92 & 65.73 & 73.03 & 31.83 & 36.80  \\

GPT-4o (w/o images)~\cite{gpt4} &  & 27.02 & 65.66 & \underline{73.99}  & 23.34 & 27.99 \\
		GPT-4o (w/o images)~\cite{gpt4} & $\checkmark$ & 21.90 & \underline{65.85} & 73.10 & 21.46 & 25.10  \\

GPT-4o~\cite{gpt4} &  & {29.37} & 65.58 & 73.86 & {24.89} & 30.19  \\
		GPT-4o~\cite{gpt4} & $\checkmark$ & 31.78 & 65.56 & {73.83}  & 31.48 & 35.90 \\

        \midrule
\rowcolor{LightPink}
		\textbf{\textsc{Zina} (Ours)} &
		$\checkmark$ & \textbf{45.15}  & \textbf{66.08} & \textbf{74.36} & \textbf{44.02} & \textbf{50.39} \\
		\bottomrule
	\end{tabular}
    \caption{Quantitative comparison with single-stage and two-stage baselines on the VisionHall dataset. \textbf{Bold} font indicates the best, and \underline{underlined} font indicates the second best. Our proposed method outperformed both single-stage and two-stage baselines in both tasks.}
	\label{table:quant_two_stage}
    \vspace{-3mm}
\end{table*}

\paragraph{Training.} 
We finetuned Qwen2.5-VL-72B-Instruct\footnote{\url{https://huggingface.co/Qwen/Qwen2.5-VL-72B-Instruct}} using LoRA \cite{lora} with a rank of $8$. The model was trained for $10$ epochs with a batch size of $4$.
We set the maximum sequence length to $2048$ tokens and the initial learning rate to $1.0 \times 10^{-4}$. We use the AdamW optimizer with $\beta_1 = 0.9$ and $\beta_2 = 0.999$, and apply cosine annealing for learning rate scheduling.

We employed LLaMA-Factory~\cite{llama-factory} to train \textsc{Zina} on the VisionHall dataset.
For training, we adopted gradient accumulation and ZeRO-3 optimization to improve memory efficiency and scalability.
All models were trained using 4 NVIDIA H200 SXM GPUs, each with 141 GB of VRAM.

\paragraph{Evaluation.} We used VLMEvalKit~\cite{vlmevalkit} to ensure reproducibility for the following models: LLaVA-1.5-7B~\cite{llava-1.5}, LLaVA-1.5-13B, Qwen2-VL-7B~\cite{qwen2vl}, LLaVA-OV-Qwen2-7B~\cite{llavaov}, LLaVA-NeXT-Qwen-32B~\cite{llavanext}, LLaVA-OV-Qwen2-72B, LLaMA-3.2-90B-Vision-Instruct~\cite{llama}, Qwen2.5-VL-72B-Instruct~\cite{qwen25vl}, and GPT-4o~\cite{gpt4}.
All experiments were reported based on a single run.

\section{Additional Experiments}
\label{sec:ablation_curation}

\subsection{Ablation on Data Curation Choices}
{
We conducted experiments to analyze the impact of data curation design choices. Specifically, we compared graph pruning strategies under two conditions: (i) deterministically pruning nodes with degree \(d < 2\), based on the average degree shown in Table~\ref{tab:vh_struct}; and (ii) randomly pruning nodes with probability \(p = 0.25\), where \(p\) denotes the probability of selecting a node for removal along with its descendants.
}

{
The results are summarized in Table~\ref{tab:curation}. We observed that the degree-based pruning strategy led to worse performance compared to the original \textsc{Zina}. This is likely due to the reduced complexity of inter-error dependencies, which results in a greater number of independent, unstructured errors.
}

{
Similarly, pruning nodes with \(p = 0.25\) yielded slightly lower performance than the original setting \((p = 0.5)\), likely because the inserted errors became slightly more complex and began to deviate from those observed in real data.
These analyses support the appropriateness of our current design choices for data curation. 
}

\subsection{Additional Quantitative Comparison}
\paragraph{Two-Stage Variants of Baselines.}
Although it is standard to compare system-level models with single-step MLLMs~\cite{unihd,fava}, we also compare \textsc{Zina} with two-stage variants of the baseline MLLMs, in addition to the original MLLMs.
These variants are modified versions of the baselines used in Table~\ref{table:quant}, converted into the same two-stage pipeline as \textsc{Zina}.

Table~\ref{table:quant_two_stage} presents the quantitative results of these two-stage comparisons.
The results show that our method even outperformed the two-stage baselines in both tasks. Some MLLMs, such as LLaVA-OV-Qwen2-7B and Qwen2.5-VL-72B-Instruct, improved performance when converted to a two-stage setup. In contrast, models such as GPT-4o and LLaVA-NeXT-Qwen-32B exhibit degraded performance in two-stages.
Together with Table~\ref{table:ablation}, these results indicate that the performance gains primarily arise from the combination of semi-synthetic data and the two-step strategy, rather than from either component in isolation.

\begin{table}[t]
	\centering
	\normalsize
    \scalebox{0.67}{
	\begin{tabular}{lccccc}
		\toprule
		\multirow{2}{*}{\textbf{Method}}
		& \multicolumn{1}{c}{\textbf{Detection}}
		& \multicolumn{2}{c}{\textbf{Editing}}
		& \multicolumn{2}{c}{\textbf{Overall}} \\
		\cmidrule(lr){2-2}\cmidrule(lr){3-4}\cmidrule(lr){5-6}
		&
		\textbf{F$_1$}
		& \textbf{CLIP-S} & \textbf{PAC-S}
		& \textbf{BERT-F$_1$} & \textbf{CLIP-F$_1$} \\
		\midrule

		\makecell[l]{LLaVA1.5-13B$^\dagger$}
            & 7.70 & 63.91 & 73.03 & 7.68 & 7.42 \\

		\makecell[l]{Qwen2.5-VL-72B$^\dagger$}
            & 24.00 & 64.95 & 73.06 & 19.95 & 24.18 \\

		\makecell[l]{LLaVA-OV-Qwen2-72B$^\dagger$}
            & 29.68 & 65.80 & 73.97 & 23.32 & 27.21 \\

        \midrule
\rowcolor{LightPink}
        \textbf{\textsc{Zina} (Ours)}
            & \textbf{45.15} & \textbf{66.08} & \textbf{74.36} & \textbf{44.02} & \textbf{50.39} \\

		\bottomrule
	\end{tabular}
    }
    \caption{Additional experiments. $^\dagger$ represents fine-tuned versions.}
	\label{table:quant_finetuned}
\end{table}

\paragraph{Finetuned Baselines.}
Table \ref{table:quant_finetuned} shows a quantitative comparison between \textsc{Zina} and the finetuned baselines.
The results show that \textsc{Zina} outperformed the finetuned baselines, indicating that the performance improvements mainly stem from the combination of synthetic data and the two-step generation strategy, rather than either component in isolation.

\subsection{Quantitative Comparison with HalLocalizer}
To evaluate the proposed method on an out-of-domain dataset, we assess \textsc{Zina} on the HalLoc~\cite{halloc} dataset. Table~\ref{table:halloc} shows that \textsc{Zina} outperformed HalLocalizer in 3 out of 4 categories. This result indicates generalization beyond VisionHall, despite the fundamental differences in dataset construction, i.e., HalLoc injects token-level errors via hallucinated QA, whereas VisionHall inserts span-level errors with a graph-based approach.

\begin{table}[H]
  \centering
\scalebox{0.8}{
  \begin{tabular}{lcccc}
    \toprule
    \textbf{Method} & \textbf{Object} & \textbf{Attribute} & \textbf{Relationship} & \textbf{Scene} \\
    \midrule
\makecell[l]{HalLocalizer$_{\textrm{best}}$}        & \underline{0.68} & \underline{0.64} & \textbf{0.71} & \underline{0.76} \\
\rowcolor{LightPink}
    \textbf{\textsc{Zina} (Ours)} & \textbf{0.82} & \textbf{0.71} & \underline{0.64} & \textbf{0.92} \\
    \bottomrule
  \end{tabular}
  }
  \caption{Results of \textsc{Zina} on HalLoc. Each column represents the F$_1$ score for the respective category.}
  \label{table:halloc}
\label{tab:halloc}
\end{table}

\section{Licenses and Intended Use}
\label{appendix:license}

\textsc{Zina} is released under the BSD 3-Clause License, and the VisionHall dataset is released under the CC BY-NC 4.0 License.
Table~\ref{tab:license} summarizes the licenses of the models and datasets used in this study.

All existing artifacts used in this study were utilized in a manner consistent with their intended use. For the artifacts we created, we define their intended use as general academic and research use, which is compatible with the original access conditions of the datasets and models employed in this study.

\section{Prompts}
\label{appendix:prompts}

\begin{table}[t]
\centering
\begin{tabular}{ll}
\toprule
\textbf{Model / Dataset} & \textbf{License} \\
\midrule
DCI dataset~\cite{dci} & CC BY-NC 4.0 \\
MHaluBench~\cite{unihd} & MIT \\
InstructBLIP~\cite{instructblip} & Research-only (non-commercial) \\
InternVL~\cite{internvl} & MIT \\
LLaVA-NeXT~\cite{llavanext} & Apache 2.0 \\
LLaVA-1.5~\cite{llava-1.5} & Apache 2.0 \\
Multimodal-GPT~\cite{mmgpt} & Apache 2.0 \\
Qwen-VL-Chat~\cite{qwen-vl} & Tongyi Qianwen \\
ShareGPT4V~\cite{sharegpt4v} & Apache 2.0 \\
BLIP-2~\cite{blip2} & BSD 3-Clause \\
GIT~\cite{git} & MIT \\
Long-CLIP~\cite{long-clip} & Apache 2.0 \\
Qwen2.5-VL~\cite{qwen25vl} & Apache 2.0 \\
BERTScore~\cite{bertscore} & MIT  \\
CLIPScore~\cite{clipscore} & MIT \\
PAC-S~\cite{pacs} & MIT \\
\midrule
\textsc{Zina} & BSD 3-Clause \\
VisionHall & CC BY-NC 4.0 \\
\bottomrule
\end{tabular}
\caption{Licenses of models and datasets used in this study.}
\label{tab:license}
\end{table}

\subsection{Baselines}

The full prompts used for baselines are described below.
These prompts are designed based on prior work \cite{fava}.
In pilot experiments, we observed that MLLMs occasionally misused the \texttt{<relation>} tag to indicate grammatical relationships rather than spatial relationships between objects, and applied the \texttt{<fact>} tag to entire sentences instead of to specific named entities.
To address these issues, we adopt \texttt{<spatial\_relation>} in place of \texttt{<relation>}, and \texttt{<named\_entities\_fact>} in place of \texttt{<fact>}.

\TColorBox{Full Prompt for Baselines}{
Given a passage with errors of image captions, identify any <spatial\_relation>, <attribute>,\\
<object>, <number>, <named\_entities\_fact>, <text> errors in the passage and add edits for errors.\\
If there are no errors, return the passage with no tags. \\
Any changes to the original passage should be marked in <> tags. \\
Below are the error definitions followed by examples of what you need to follow.\\
Definitions:\\
\\
**Types of Errors and Definitions:**\\
1. **<spatial\_relation> (Spatial Relation Error)**: Incorrect spatial relationship between entities or concepts. (Only those involving spatial relationships are permitted, not grammatical relationship errors.)\\
   - Example: "an apple on the table" → "an apple <spatial\_relation>under</spatial\_relation> the table"\\
\\
2. **<attribute> (Attribute Error)**: Incorrect attributes such as adjectives or adverbs describing objects or concepts.\\
   - Example: "Red sky" → "<attribute>Blue</attribute> sky"\\
\\
3. **<object> (Object Error)**: Incorrect specific objects or entities.\\
   - Example: "There is a chair on the table." → "There is a <object>book</object> on the table."\\
\\
4. **<number> (Number Error)**: Incorrect quantities or numerical values.\\
   - Example: "3 cats" → "<number>5</number> cats"\\
\\
5. **<named\_entities\_fact> (Factual Error)**: Error related to named entities. This tag should not be applied to anything other than named entities. For example, "John F. Kennedy Center" is acceptable, but "center" is not.\\
   - Example: "The image is of the John F. Kennedy Center." → "The image is of the <named\_entities\_fact>white house</named\_entities\_fact>."\\
\\
6. **<text> (Text Error)**: Incorrect scene texts (text that appears in an image).\\
   - Example: "A car is parked under a sign that says 'Restaurant'." → "A car is parked under a sign that says <text>'Hotel'</text>."\\
\\
\#\# Example\\
Passage: [few-shot-example]  \\
\\
Reference: [few-shot-example] \\
\\
Edited: [few-shot-example] \\
\\
\#\# Task\\
\\
Now detect errors and include edits in the following passage like done in the example above.\\
Include error tags <> for ANYTHING YOU CHANGE IN THE ORIGINAL PASSAGE.\\
\\
Passage: [Original]\\
\\
Reference: [Reference] \\
\\
Edited:
}

\subsection{Detection and Reviewer MLLMs}

The full prompts used for the detection MLLM $\mathcal{M}_\mathrm{det}$ and the reviewer MLLM $\mathcal{M}_\mathrm{rev}$ are described below.
These prompts are also designed based on prior work \cite{fava}.
\TColorBox{Full Prompt for $\mathcal{M}_\mathrm{det}$}{
You are given an image, a generated caption (Original), and a human reference caption (Reference).\\  
 Your task is to detect any single word in the Original that is NOT supported by the image\\  
 (hallucinations) and label each of them with exactly one tag from:\\  
 \\  
   • object              – wrong or missing object\\
   • spatial\_relation    – wrong spatial or positional term\\  
   • attribute           – wrong adjective or adverb\\  
   • number              – wrong quantity\\  
   • text                – incorrect visible text\\  
   • named\_entities\_fact – incorrect named entity\\  
 \\  
 Return ONLY a comma‑and‑slash separated list of “word, tag” pairs, e.g.  \\  
   three, number / apples, object  \\  
 If no hallucinations exist, return exactly:  \\  
   none\\  
 \\  
 ----------------  FEW‑SHOT EXAMPLES  ----------------\\  
 \\  
 \# 1. number\\  
 Original : "There are three cats."\\  
 Reference: "Two cats are on the sofa."\\  
 Output   : three, number\\  
 \\  
 \# 2. spatial\_relation\\  
 Original : "An apple on the table."\\  
 Reference: "An apple is under the table."\\  
 Output   : on, spatial\_relation\\  
 \\  
 \# 3. attribute\\  
 Original : "Red sky over the mountains."\\  
 Reference: "Blue sky over the mountains."\\  
 Output   : Red, attribute\\  
 \\  
 \# 4. object\\  
 Original : "There is a chair on the table."\\  
 Reference: "There is a book on the table."\\  
 Output   : chair, object\\  
 \\  
 \# 5. named\_entities\_fact\\  
 Original : "The image shows the John F. Kennedy Center."\\  
 Reference: "The image shows the White House."\\  
 Output   : John F. Kennedy Center, named\_entities\_fact\\  
 \\  
 \# 6. text\\  
 Original : "A sign says 'Restaurant'."\\  
 Reference: "A sign says 'Hotel'."\\  
 Output   : Restaurant, text\\  
 \\  
 ------------  NOW PROCESS THIS SAMPLE  ------------\\  
 \\  
 Original:\\  
 [Original]\\  
 \\  
 Reference:\\  
 [Reference]\\  
 \\  
 Output:\\  
 
}

\TColorBox{Full Prompt for $\mathcal{M}_\mathrm{rev}$}{
You are given an image, the same reference caption, and an “Original” caption\\  
 in which candidate hallucination words have been wrapped with XML tags:\\  
 \\  
   <object> … </object>\\  
   <spatial\_relation> … </spatial\_relation>\\  
   <attribute> … </attribute>\\  
   <number> … </number>\\  
   <text> … </text>\\  
   <named\_entities\_fact> … </named\_entities\_fact>\\  
 \\  
 For EACH tagged word decide:\\  
   – If it must be corrected, return the corrected word.\\  
   – If it is already correct, return the original word unchanged.\\  
 \\  
 Return ONLY a result “tagged\_segment: replacement”, e.g.\\  
   <object>chair</object>: book\\  
 \\  
 ----------------  FEW‑SHOT EXAMPLES  ----------------\\  
 \\  
 \# 1. number: wrong or missing object\\  
 Original : There are <number>three</number> cats.  \\  
 Reference: Two cats are on the sofa.  \\  
 Output   : <number>three</number>: two\\  
 \\  
 \# 2. spatial\_relation: wrong spatial or positional term\\  
 Original : An apple <spatial\_relation>on</spatial\_relation> the table.  \\  
 Reference: An apple is under the table.  \\  
 Output   : <spatial\_relation>on</spatial\_relation>: under\\  
 \\  
 \# 3. attribute: wrong adjective or adverb\\  
 Original : <attribute>Red</attribute> sky over the mountains.  \\  
 Reference: Blue sky over the mountains.  \\  
 Output   : <attribute>Red</attribute>: Blue\\  
 \\  
 \# 4. object: wrong object\\  
 Original : There is a <object>chair</object> on the table.  \\  
 Reference: There is a book on the table.  \\  
 Output   : <object>chair</object>: book\\  
 \\  
 \# 5. named\_entities\_fact: incorrect named entity \\  
 Original : The image shows the <named\_entities\_fact>John F. Kennedy Center</named\_entities\_fact>.  \\  
 Reference: The image shows the White House.  \\  
 Output   : <named\_entities\_fact>John F. Kennedy Center</named\_entities\_fact>: White House\\  
 \\  
 \# 6. text: incorrect visible text\\  
 Original : A sign says <text>'Restaurant'</text>.  \\  
 Reference: A sign says 'Hotel'.  \\  
 Output   : <text>'Restaurant'</text>: 'Hotel'\\  
 \\  
 ------------  NOW PROCESS THIS SAMPLE  ------------\\  
 \\  
 Original:  \\  
 [Original]\\  
 \\  
 Reference:  \\  
 [Reference]\\  
 \\  
 Instructions:  \\  
 - Only look at segments already wrapped in XML tags in the Original (`<object>…</object>`, `<spatial\_relation>…</spatial\_relation>`, etc.).  \\  
 - Do **not** add any new tags.  \\  
 - Decide for **each** existing tag whether it needs correction, but then choose **only one** tagged segment to report (the first or most obvious error).  \\  
 - Output **exactly one** line in the form:  \\  
   `<tag>word</tag>: corrected\_word`  \\  
 —and nothing else.\\  
 
}
\subsection{Synthetic Training Data Curation}

The full prompts used for the Error Insertion (EI) module  are shown below:
\WideTColorBox{Full Prompt for Inserting Errors}{
**Objective:**  
 \\Insert six types of errors (as defined below) into the provided candidate sentence (cand) while following the guidelines. Use the provided references (refs) as needed.
 \\
 \\**Instructions:**
 \\
 \\- **Modify Only Existing Text:**
 \\  - Only change words or phrases that already appear in the candidate sentence (cand).
 \\  - Do not add any extra words, phrases, or information that are not in cand.
 \\  - The overall content must remain exactly as in the original cand aside from the inserted error tags.
 \\
 \\- **Maintain Consistency:**
 \\  - When replacing a word, update every reference to that word in the sentence to preserve consistency.  
 \\    - *Example:* Changing “three apples” to “seven apples” requires updating all mentions of “three apples” in the sentence.
 \\  - If you replace one term (e.g., “train station” with “bus stop”), ensure that all instances of “train station” are changed accordingly.
 \\
 \\- **Scene Text Errors:**
 \\  - Insert text errors only when there is an existing reference to scene text in the original sentence.
 \\
 \\- **Error Annotation Format:**
 \\  - Use the following XML-like format to mark errors:
 \\    - `<object original="original object name" id="E1">wrong object name</object>`
 \\    - `<attribute original="original attribute" id="E2" parent="E1">wrong attribute</attribute>`
 \\  - **Rules for Error Tags:**
 \\    - Each error must include:
 \\      - The type of error.
 \\      - A sequential ID.
 \\      - The original text.
 \\      - A parent dependency (if applicable).
 \\        - Once an object is mislabeled, all subsequent references to it must consistently reflect the same error. 
 \\        - If subsequent text depends on an existing error, use the same parent ID to denote this dependency. 
 \\    - Do not nest error tags.
 \\    - Do not use the same ID for both a current element and its parent.
 \\    - After applying modifications, ensure there are no duplicate IDs.
 \\    - Modifications must be applied to whole words only (do not split a word into parts). For example, changing "beach" by replacing "bea" with "city" and "ch" with "desert" is not allowed.
 \\    - Do not insert tags that result in semantically equivalent replacements. For example, changing "a" to "<number original="a" id="E1">one</number>" is not allowed.
 \\
 \\- **Types of Errors:**
 \\
 \\  1. **spatial\_relation (Spatial Relation Error):**  
 \\     - Modify only prepositions that indicate spatial relationships (no grammatical errors allowed).  
 \\     - *Example:*  
 \\       `"an apple on the table"` →  
 \\       `"an apple <spatial\_relation original="on" id="E1">under</spatial\_relation> the table"`
 \\
 \\  2. **attribute (Attribute Error):**  
 \\     - Change descriptive attributes such as adjectives or adverbs.  
 \\     - *Example:*  
 \\       `"Red sky"` →  
 \\       `"<attribute original="Red" id="E1">Blue</attribute> sky"`
 \\
 \\  3. **object (Object Error):**  
 \\     - Replace names of generic objects or entities that are common nouns (i.e., not proper names).
 \\     - *Example:*  
 \\       `"There is a chair on the table."` →  
 \\       `"There is a <object original="chair" id="E1">book</object> on the table."`
 \\    - Note: For any proper noun or official name (e.g., "John F. Kennedy Center" or "Fuji"), use the fact error as described next.
 \\
 \\  4. **fact (Factual Error):**  
 \\     - Replace proper, named entities with incorrect names. Apply this error type only to entities that are proper names (e.g., formal institution names, landmarks, or official titles).
 \\     - *Example:*  
 \\       `"The image is of the John F. Kennedy Center."` →  
 \\       `"The image is of the <named\_entities\_fact original="John F. Kennedy Center" id="E1">white house</named\_entities\_fact>."`
 \\     - Important: For any non-proper, generic object or entity (even if it might appear to be an entity), use the object error type instead.
 \\
 \\  5. **number (Number Error):**  
 \\     - Change numerical values or quantities.  
 \\      - *Example:*  
 \\       `"3 cats"` →  
 \\       `"<number original="3" id="E1">5</number> cats"`
 \\    - If you change a singular quantity to plural (or vice versa), make sure the associated noun reflects the correct number (i.e., use plural form for plural numbers and singular form for singular numbers). No annotation tags are needed for singular/plural noun form adjustments.
 \\        - *Example:*  
 \\        `"a cat"` →  
 \\        `"<number original="a" id="E1">5</number> cats"`
 \\
 \\  6. **text (Text Error):**  
 \\     - Change scene text (i.e., text that appears within the image).  
 \\     - *Example:*  
 \\       `"A car is parked under a sign that says 'Restaurant'."` →  
 \\       `"A car is parked under a sign that says <text original="'Restaurant'" id="E1">'Hotel'</text>."`
 \\
 \\- **Output Format:**
 \\  - Provide the modified sentence along with error annotations in the following JSON format:
 \\  
 \\    ```json
 \\    \{
 \\      "original": "Two apples in front of a cat. The apples are Fuji.",
 \\      "modified": "<number original="Two" id="E1">nine</number> <object original="apples" id="E2">bananas</object> <spatial\_relation original="in front of" id="E3">on</spatial\_relation> <object original="a cat" id="E4">a table</object>. The <object original="apples" id="E5" parent="E2">bananas</object> are <named\_entities\_fact original="Fuji" id="E6">Cavendish</named\_entities\_fact>."
 \\    \}
 \\    ```
 \\  
 \\  - **Important:**  
 \\    - Only modify words that exist in the original cand.
 \\    - Do not add any additional words or phrases.
 \\
 \\- **Input:**
 \\  - **Original Sentence (cand):**  
 \\    {cand}
 \\  - **Reference (refs):**  
 \\    {refs}
}

\WideTColorBox{Full Prompt for Checking and Revising}{
You are a quality control assistant tasked with evaluating the output `generated\_xml` produced by the LLM (which is the modified version of the provided `original` candidate sentence). Your objective is to ensure that `generated\_xml` fully complies with all specified guidelines. Please review the output using the checklist below and then provide a JSON-formatted report detailing your findings. If modifications are necessary, include the revised output.
 \\
 \\**Checklist:**
 \\
 \\1. **Original Sentence Integrity:**
 \\   - Verify that the value of the `"original"` key exactly matches the provided candidate sentence.
 \\   - Ensure that no modifications, additions, or extra words/phrases have been introduced in the `"original"` value.
 \\
 \\2. **Modified Sentence Requirements:**
 \\   - Confirm that the `"modified"` value is the candidate sentence with error annotations inserted.
 \\   - Check that only words or phrases from the candidate sentence have been modified—no new content should be added.
 \\   - *Note:* The meaning of the `"modified"` value may differ from the original.
 \\
 \\3. **Error Annotation Types:**
 \\   - Confirm that only the following six error types are used: **spatial\_relation, attribute, object, number, named\_entities\_fact, text**.
 \\   - Verify that each error annotation exactly matches one of these types and is applied only in the appropriate context (for example, text errors are used only for scene text).
 \\
 \\4. **Error Tag Format and Consistency:**
 \\   - Each error annotation must use the XML-like tag format:
 \\     ```xml
 \\     <error\_type original="original text" id="E\#">modified text</error\_type>
 \\     ```
 \\   - Ensure that:
 \\     - The tag name is exactly one of the specified error types.
 \\     - The `original` attribute contains the exact original word or phrase from the candidate sentence.
 \\     - The `id` attribute is sequential (e.g., E1, E2, …) with no duplicates.
 \\     - For dependent errors, the child error tag must include the correct `parent` attribute referencing the appropriate error ID.
 \\   - Error tags should only be applied to whole words (no splitting of words), and no error tag should be nested within another.
 \\   - **Whitespace Requirement:** Since tags are treated like words, ensure there is a space immediately after a closing tag. For example, `<attribute>blue</attribute> <object>book</object>` is correct, whereas `<attribute>blue</attribute><object>book</object>` is not acceptable.
 \\
 \\5. **Consistency in Modifications:**
 \\   - Verify that if a word is modified in one location, every occurrence of that word in the sentence is modified consistently.
 \\   - Ensure that the same error tag (and, if applicable, the same `parent` attribute) is used for all repeated instances of that word.
 \\   - When one modification depends on another error, ensure that a proper tree structure is maintained using the `parent` attribute.
 \\
 \\6. **No Semantically Equivalent Replacements:**
 \\   - Ensure that error annotations do not result in semantically equivalent replacements (e.g., do not replace “a” with a number tag representing “one”).
 \\
 \\7. **Singular/Plural Adjustments:**
 \\   - If changing a singular quantity to plural (or vice versa), ensure that the associated noun reflects the correct number (i.e., use plural form for plural numbers and singular form for singular numbers).
 \\   - When modifying the noun’s number, the change must be performed via a number error tag applied to the noun, with the number error tag for the quantity serving as the parent. If the noun is already modified by an object (or another) tag, the existing tag takes priority but the `parent` attribute should be added.
 \\     - **Example:**
 \\       - `a cat` → `<number original="a" id="E1">three</number> <number original="cat" id="E2" parent="E1">cats</number>`
 \\       - `a cat` → `<number original="a" id="E1">three</number> <object original="cat" id="E2" parent="E1">dogs</object>`
 \\
 \\8. **No Article Swapping:**  
 \\   - Modifications that swap articles (e.g., changing "a" to "the" or vice versa) are not permitted.
 \\
 \\9. **No Unjustified Modifications:**  
 \\   - Do not allow changes that cannot be determined solely from the original word. For example, altering `two bible 'Guide'` to `two <named\_entities\_fact>quran</named\_entities\_fact> <text>bible</text>` is unacceptable, as it is unrealistic to deduce the appropriate modification for "bible" from the candidate sentence alone.
 \\
 \\10. **No Matching Words Between Tagged and Untagged Text:**  
 \\   - Ensure that the words within the error tags do not exactly match the corresponding words in the original, untagged text. For instance, transforming "To the left" into `<spatial\_relation>To the right</spatial\_relation>` is incorrect because the words "To" and "the" are repeated. The proper transformation is to only enclose the modified word: "To the left" should become "To the <spatial\_relation>right</spatial\_relation>".
 \\
 \\11. **Application of Error Annotations for Generic Objects or Entities:**
 \\  - Fundamental Principle: For generic objects or entities expressed by common nouns (i.e., non-proper nouns), do not use the named\_entities\_fact tag. Always use the object tag.
 \\
 \\  - named\_entities\_fact (Factual Error): This tag should only be used for proper names such as formal institution names, famous landmarks, or official titles. It is only applicable when an incorrect proper name needs to replace the original.
 \\
 \\  - object (Object Error): For common objects or entities (non-proper nouns), use this tag instead. In cases where the word does not represent a proper noun, the object tag must be used rather than named\_entities\_fact.
 \\
 \\---
 \\
 \\**Output Format:**
 \\
 \\After your evaluation, please produce a JSON object with the following keys:
 \\
 \\- `"need\_update"`: A boolean indicating whether the output requires modifications.
 \\- `"updated\_xml"`: The revised output (if modifications are needed) or a copy of the compliant output.
 \\- `"reason"`: A detailed explanation listing any detected violations with clear references to the specific rule(s) broken, or a statement confirming full compliance if no issues are found.
 \\
 \\**Example:**
 \\- `original`: a red apple
 \\- `generated\_xml`: "<number original="a">3</number> <attribute original="red" >blue</attribute> <object original="apple">books</object>"
 \\
 \\```json
 \\\{
 \\  "need\_update": true,
 \\  "updated\_xml": "<number original="a" id="E1">3</number> <attribute original="red" id="E2">blue</attribute> <object original="apple" id="E3" parent="E1">books</object>",
 \\  "reason": "yyy"
 \\\}
 \\```
 \\
 \\**Input Data:**
 \\- **`original`:**  
 \\  \{original\}
 \\- **`generated\_xml`:**  
 \\  \{output\}
}

\end{document}